\documentclass[lettersize,journal]{IEEEtran}

\ifCLASSINFOpdf
\else
\fi
\usepackage{hyperref}	
\usepackage[table,xcdraw]{xcolor}
\usepackage[inkscapelatex=false]{svg}
\usepackage{graphicx} 
\usepackage{float} 
\usepackage{epstopdf}
\usepackage{caption}
\usepackage{booktabs}
\usepackage[export]{adjustbox} 
\usepackage{multirow}
\usepackage{amssymb}
\begin{document}
%
\title{Incorporating Long-term Data in Training Short-term Traffic Prediction Model}


\author{Xiannan Huang~\IEEEmembership{},
Shuhan Qiu~\IEEEmembership{},
Yan Cheng~\IEEEmembership{},
Quan Yuan~\IEEEmembership{}, and
Chao Yang~\IEEEmembership{}
\thanks{Corresponding authors: Chao Yang.}
\thanks{Xiannan Huang, Shuahan Qiu and Yan Cheng are with the Key Laboratory of Road and Traffic Engineering, Ministry of Education at Tongji University, 4800 Cao’an Road, Shanghai, 201804, China (e-main: huang\_xn@tongji.edu.cn; qiusuan@tongji.edu.cn; yan\_cheng@tongji.edu.cn)}

\thanks{Quan Yuan is with the Urban Mobility Institute, Tongji University, 1239 Siping Road, Shanghai, 200082, China (e-mail:quanyuan@tongji.edu.cn)}

\thanks{
Chao Yang is with the Key Laboratory of Road and Traffic Engineering, Ministry of Education at Tongji University, 4800 Cao’an Road, Shanghai, 201804, China, also with the Urban Mobility Institute, Tongji University, 1239 Siping Road, Shanghai, 200082, China (e-mail: tongjiyc@tongji.edu)
}
}

%



\IEEEtitleabstractindextext{%
\begin{abstract}
Short-term traffic volume prediction is crucial for intelligent transportation system and there are many researches focusing on this field. However, most of these existing researches concentrated on refining model architecture and ignored amount of training data. Therefore, there remains a noticeable gap in thoroughly exploring the effect of augmented dataset, especially extensive historical data in training. In this research, two datasets containing taxi and bike usage spanning over eight years in New York were used to test such effects. Experiments were conducted to assess the precision of models trained with data in the most recent 12, 24, 48, and 96 months. It was found that the training set encompassing 96 months, at times, resulted in diminished accuracy, which might be owing to disparities between historical traffic patterns and present ones. An analysis was subsequently undertaken to discern potential sources of inconsistent patterns, which may include both covariate shift and concept shift. To address these shifts, we proposed an innovative approach that aligns covariate distributions using a weighting scheme to manage covariate shift, coupled with an environment aware learning method to tackle the concept shift. Experiments based on real word datasets demonstrate the effectiveness of our method which can significantly decrease testing errors and ensure an improvement in accuracy when training with large-scale historical data. As far as we know, this work is the first attempt to assess the impact of contiguously expanding training dataset on the accuracy of traffic prediction models. Besides, our training method is able to be incorporated into most existing short-term traffic prediction models and make them more suitable for long term historical training dataset.
\end{abstract}

}

\maketitle

\IEEEdisplaynontitleabstractindextext

%
\IEEEpeerreviewmaketitle

\section{Introduction}
%
%
%
%
\IEEEPARstart{T}{he} problem of short-term traffic prediction means forecasting the traffic volume in given regions for a future period and it plays an important role in intelligent transportation system. Besides, the primary focus about this topic currently revolves around refining model architectures to enhance predictive accuracy on the test set \cite{Zou2024MTSTNetAN,Wei2024,Yan2024ProSTformerPS}. However, recent advancements in machine learning domains, such as computer vision and natural language processing, exemplified by works like CLIP \cite{Radford2021LearningTV} and GPT \cite{Radford2018ImprovingLU}, underscore the necessity for both larger datasets and superior model architectures to achieve better accuracy. Consequently, to achieve superior precision in short-term traffic prediction, a larger dataset might be imperative. Solely concentrating on enhancing model structures may be insufficient.  

There are a few works using multiple spatial-temporal datasets to train a foundation model for traffic prediction \cite{Li2024OpenCityOS,10.1145/3637528.3671662,li2024UrbanGPT}. These researches used different kinds of datasets such as traffic flow, taxi demand, bike demand, traffic speed  \cite{Li2024OpenCityOS,10.1145/3637528.3671662,li2024UrbanGPT}, dynamic population \cite{10.1145/3637528.3671662}, and even some datasets not part of traffic domain such as cellular network usage \cite{10.1145/3637528.3671662} and crimes \cite{li2024UrbanGPT}, and these datasets are from different cities such as New York, Chicago, Beijing etc. However, for a specific prediction task, even though incorporating different types of datasets might be valuable, there exist a more straightforward way to augment training dataset, which is simply concluding the traffic data in earlier time. This is because in the actual operation of the transportation system, a large amount of historical data will accumulate over time, and even close to ten years of historical data can be obtained. This historical data is from the same city and the same traffic mode as the prediction task, which might be more informative than the different types of data from different cities. 

However, current researches almost only use the data of the last few months \cite{Huang2024MTLMetroAD,Han2024KnowledgeBasedMR} to one \cite{Wei2024} or two years \cite{Yan2024ProSTformerPS}, the reason may be that, the patterns of data that are too old are inconsistent with the current patterns \cite{Wang2024EvaluatingTG}, which can be called distribution shift in machine learning domain, and we will analysis this inconsistency later. Even though such disadvantages exist, just excluding so many history records could be a waste of data. Therefore, analyzing the effect of including more historical data and designing method to make better use of these data is necessary.

It is noticed that there are some methods addressing distribution shift in general time series prediction problem. Some works refer to the theorems in transfer learning and use specially designed modules to align the distributions of covariates \cite{Kim2022ReversibleIN} or features \cite{Du2021AdaRNNAL} in different periods. And other works proposed to adjust some parameters to fit the new patterns according to the new data \cite{You2021LearningTL,zhang2023onenet}. These works designed innovative methods to address different distribution shift problems and they are all insightful. However, the traffic prediction problem shows some distinctive characteristics compared to the general time series prediction problem, so it is needed to design specific module to handle distribution shift in this problem. We summarize the distinctive feature as follows:

First, traffic prediction is a multi-variable prediction task, with each variable related to the traffic volumes in a specific region. And the change of patterns in different regions may be different. In another word, it is possible that the pattern of one variable changed but the patterns of the other variables remain unchanged. For example, if a new supermarket opens in a specific region, then the traffic patterns in this region will change but the patterns in other regions will not change significantly. Therefore, the distribution shift of different variables may be different, which is distinct from the general time series prediction problem.

Second, it is needed to consider the correlation between different regions in traffic prediction task. But the spatial relationship may change over time. For example, some paper proposed that the relationship between different regions could beyond spatial adjacency \cite{Liu2022SpatialTemporalDG,Liu2024STDAGCNAS}, for example, the regions of similar land usage type should be connected \cite{Wei2024}. Therefore, if the land usage type in one region change, there might be some new spatial connections between this region and other regions and some old connections could diminish. Besides, if a new road opens, the relationship of different regions might change too. Therefore, the changing relationship between regions is also an important difference between traffic prediction problem and general time series prediction problem.

Therefore, even though we can draw some inspiration from the works addressing distribution shift problem in general time series prediction task, modules specifically designed to address the distinctive features in traffic prediction task are needed. 

According to the analysis above, we first collected two dataset containing bike and taxi usage in a period over 8 years in New York City and used the data in most recent 12,24,48 and 96 months to train some short-term traffic prediction models. Then the prediction accuracy in the test set was observed. It was founded that training with larger dataset cannot ensure better prediction accuracy. Afterwards, we designed an innovative method which can help these traditional short-term traffic prediction models capture the changing traffic patterns better and can be easily incorporated into most traditional short-term traffic prediction models. This method consists of different modules for concept shift and covariate shift and can make the larger dataset lead to the better accuracy. We summarize our main contribution as follows:
\begin{enumerate}
    \item We analyzed the effect of incorporating more historical data in training short-term traffic prediction models and found that more historical data could cause the decrease in prediction accuracy sometimes.
    \item We analyzed distribution shift patterns in short-term traffic prediction problem and proposed a training method to address the covariate shift and concept shift concurrently. 
    \item The experiments in real world datasets demonstrate the effectiveness of our method, which can lead to an improvement in MAE of 2\% to 5\% and ensure that larger datasets contribute to better accuracy.
\end{enumerate}

\section{Literature Review}
\subsection{Short-term traffic prediction models}
The problem of short-term traffic prediction involves forecasting the traffic volume in given regions for a future period. This challenge is characterized by two dimensions: time and space. Specifically, the traffic volume at different time slices forms a time series for a given region. Besides, for a given time slice, data from different spatial locations constitute a graph. Consequently, modeling short-term traffic prediction necessitates considerations for temporal and spatial aspects. Prior research commonly leverages convolution, attention, or recurrent neural networks to model temporal relationships \cite{Yu2017SpatiotemporalRC}, while spatial relationships are often addressed using spatial convolution or spatial attention networks \cite{Zheng2019GMANAG,Guo2019AttentionBS}.

In terms of overall model structure, three main approaches are identified. The first integrates time and space models into a unified spatiotemporal module, where input data passes through multiple spatiotemporal modules, extracting intricate spatiotemporal information for subsequent prediction \cite{Guo2019AttentionBS}. The second incorporates spatial modules into the temporal module, for instance, using graph convolution to replace fully connected layers in recurrent neural networks \cite{Li2017DiffusionCR}. The third model structure involves applying a spatial module to data in each time step, extracting spatial information for each time slice, and feeding it into the temporal module to obtain predictions \cite{Zhao2018TGCNAT}.

Moreover, certain studies aim to jointly explore spatiotemporal information in a single module rather than through separate processing \cite{Song2020SpatialTemporalSG}. Additionally, specialized approaches such as self-supervised learning \cite{Shao2022PretrainingES,Ji2022SpatioTemporalSL}, or meta-learning \cite{Pan2019UrbanTP} have been proposed to enhance model training effectiveness. For further insights on this topic, interested readers can refer to these recent review articles \cite{Luo2023STG4TrafficAS,Hou2022DeepLM,Jin2023SpatioTemporalGN}.

As for training spatial-temporal model with larger dataset, there are some works \cite{Li2024OpenCityOS,10.1145/3637528.3671662,li2024UrbanGPT} using many different kinds of datasets in different cities to calibrate prediction models. But these works are mainly inspired by the larger language models and the focuses of their researches are versatility, zero-shot ability and scaling law. Therefore, the out-of-distribution problem has not been carefully analyzed in these papers. 

There are also some works focusing on distribution shift in spatial-temporal prediction task. \cite{Wang2024EvaluatingTG} built some datasets spinning two years and used the data in the first year to train models and test the models using data from the later year. It was observed that the prediction error could increase significantly when testing with out-of-distribution samples. \cite{Deciphering2023} used causal inference method to ameliorate the negative effect caused by distribution shift. \cite{Maintaining2023} proposed that there are stable patterns in different periods and design some delicate method to learn these invariant patterns from data. 

All the works above are inspiring, however, as highlighted in the introduction, most of these works do not focus on the effect of adding more historical data into training set and device method to make better use of these data. 
\subsection{Distribution Shift in Time Series Prediction}

In the realm of time series data forecasting, methods for addressing distributional shift can be primarily categorized as follows:

The first category employs alignment methods. For example, \cite{Kim2022ReversibleIN} utilized normalization parameters specifically to different time periods to align the mean and variance of the data and \cite{Liu2022NonstationaryTE} refined the structure of transformer to make it more suitable for the aligned data. \cite{Fan2023DishTSAG} employed neural networks to predict the mean and variance of future data, facilitating adaptive normalization based on the statistical characteristics of oncoming data. \cite{Du2021AdaRNNAL} sliced the time series, employing neural networks to extract features and aligning features in each slice. While these approaches are insightful, alignment methods are generally more suitable for addressing covariate shift issues, and addressing different patterns can be challenging.

The second category of addressing distribution shift involves updating parameters. 
\cite{Bifet2007LearningFT} suggested updating the model parameters directly, and \cite{Li2022DDGDADD} proposed a method involving predicting the distribution of future data, generating future data according to these distributions, and training the model using the generated data. \cite{Arik2022SelfAdaptiveFF} used some self-supervised learning tasks in test period to refresh some parameters to make the model be able to address the distribution mismatch between training dataset and testing dataset. Additionally, \cite{zhang2023onenet} proposed a model ensemble approach, continuously updating the weights of each model over time to adapt to varying distributions of future data. \cite{You2021LearningTL} enhanced the model's ability to generalize to unknown data through meta-learning updates. \cite{Cai2023MemDAFU} proposed a method involving updating only a small set of parameters and incorporating a memory database which allows the model to reference the most similar memories and adapt to future changes.

However, these methods are all based on general time series prediction problem and do not propose special modules for the distinctive patterns in traffic prediction task.

\section{PRE-EXPERIMENT}
To observe the changes in prediction error when training sets contain more history data, some pre-experiments were conducted. And we will elaborate these pre-experiments as follows.
\subsection{Data}
\begin{figure}[!h]
    \centering
    \includegraphics[width=0.95\linewidth]{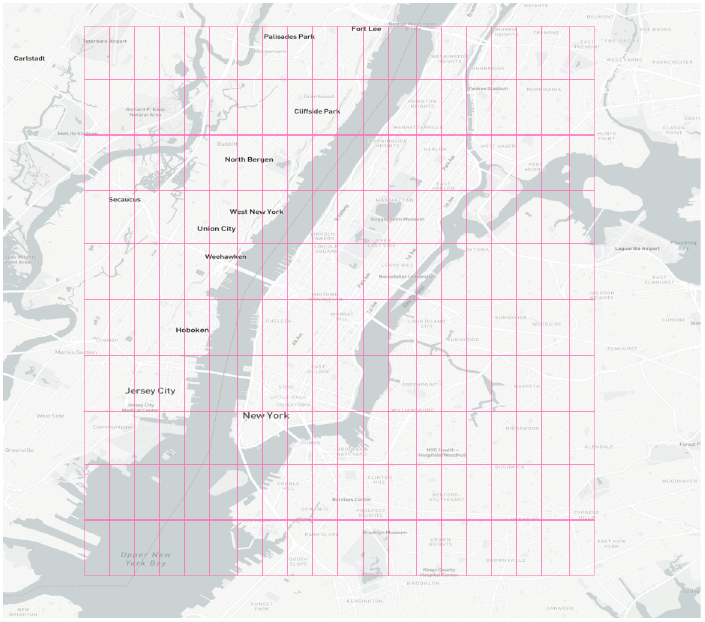}
    \caption{The girds for NYCBIKE dataset}
    \label{fig:bike grid}
\end{figure}

\begin{figure}[!h]
    \centering
    \includegraphics[width=\linewidth]{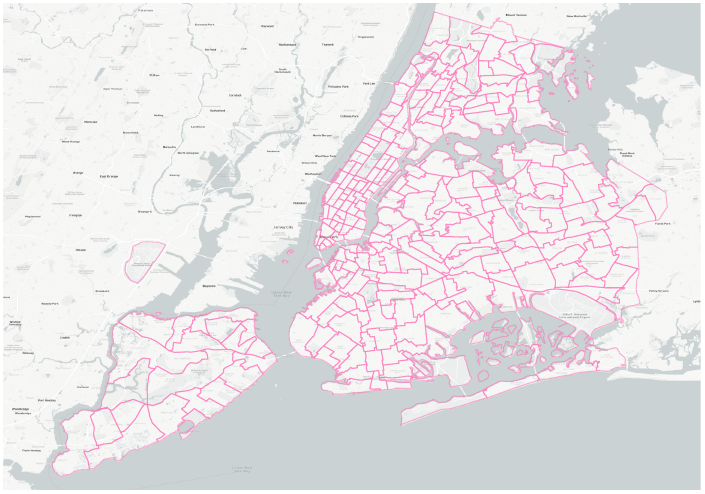}
    \caption{The regions for NYCTAXI dataset}
    \label{fig:taxi region}
\end{figure}

Two datasets spanning over eight years were collected. One comprises bike trip data in New York City from January 2015 to July 2023 \footnote{https://citibikenyc.com/system-data}, which will be called “NYCBIKE” in the following part, while the other dataset consists of yellow taxi trip data in New York City during the same period \footnote{https://www.nyc.gov/site/tlc/about/tlc-trip-record-data.page}, which will be called “NYCTAXI” in the following part. For the NYCBIKE dataset, New York city was divided into two hundred grids. And the hourly bike pickup and drop-off counts for each grid can be obtained utilizing the latitude and longitude information of the origin and destination points provided by the original website. The grid partitioning scheme follows \cite{LibCity}, and each grid has a width of 0.759km and a length of 2.224km. Besides, the hour bike usage numbers in some grids are small, so we deleted the grids where the average bike usage is less than 10, resulting 114 girds left. Regarding the taxi dataset, as the order data directly indicates the departure and arrival regions for each trip, the hourly taxi departure and arrival counts for each region could be extracted. The regional partitioning scheme is based on the official division provided by the taxi dataset website. There are 264 regions in total, and the average area of these regions is $5.250km^2$. Also, we delete the regions where the average taxi usage number is less than 10 which resulting in 74 regions left. We plot the region for bike and taxi dataset in the Figure \ref{fig:bike grid} and Figure \ref{fig:taxi region}.
\subsection{Prediction Task}

Our task involves predicting the traffic volume in each region for the next hour based on the traffic volume of the preceding six hours. The test set is defined as the data in July 2023. In other words, our aim is to calibrate a model using data before July 2023 and subsequently forecast the traffic volume in July 2023. Four training scenarios are considered: the first utilizes the previous 12 months for training, encompassing June 2023 to June 2022; the second employs the previous 24 months of data for training, covering June 2021 to June 2022; the third and fourth scenarios involve using the previous 48 months and 96 months of data as training sets, respectively. For each training scenario, the data for in last month is set aside as a validation set to determine whether training should be terminated prematurely. 

\subsection{Baseline Models and Evaluation Metrics}
Five classical and commonly used spatiotemporal prediction models were employed as baseline models. 
\begin{enumerate}
    \item \textbf{STGCN} \cite{Yu2017SpatiotemporalGC}: This method encapsulates temporal convolution and spatial convolution into a spatiotemporal block. Multiple spatiotemporal blocks are employed to integrate spatiotemporal information, ultimately yielding the final predicted values.
    \item \textbf{MTGNN} \cite{Wu2020ConnectingTD}: A learnable node embedding is initially constructed, and the adjacency matrix between nodes is adaptively formed based on the similarity of node embeddings. Subsequently, a series of iterations involve temporal convolution and graph convolution are conducted. This process integrates information from both the temporal and spatial axes, ultimately resulting in the prediction values.
    \item \textbf{DCRNN} \cite{Li2017DiffusionCR} :This model primarily employs diffusion graph convolution modules to replace the fully connected modules in traditional Recurrent Neural Networks (RNNs). This transformation converts the conventional RNN into an RNN capable of considering spatial relationships, thereby modeling spatiotemporal dependencies.
    \item \textbf{GWNET} (Graph Wavenet) \cite{Wu2019GraphWF}: Adaptive adjacency matrices are employed in this model to consider spatial relationships. Additionally, causal dilated temporal convolution and spatial convolution are utilized to model the temporal and spatial dependencies. The modules for temporal and spatial convolution are encapsulated into a spatiotemporal block. Multiple spatiotemporal blocks concurrently explore different levels of spatiotemporal information, ultimately yielding the final predicted values.
    \item \textbf{AGCRN} \cite{Bai2020AdaptiveGC}: The approaches is highly similar to DCRNN. It employs graph convolution to replace the fully connected modules in an RNN. In comparison to DCRNN, the AGCRN model introduces an adaptive adjacency matrix to account for spatial relationships. Additionally, it allows for node-specific parameters in the graph convolution layer, enabling the convolution parameters for each node to be adjustable.
\end{enumerate}

The evaluation metrics for forecasting accuracy in our study are the Mean Absolute Error (MAE) and Root Mean Squared Error (RMSE). The formulations for MAE and RMSE are defined as Eq.\ref{mae} and Eq.\ref{rmse}:
\begin{equation}
      MAE=\frac{1}{n}\sum_{i=1}^{n}\left|y_{pred_i}-y_{actual_i}\right|
      \label{mae}
\end{equation}
\begin{equation}
   RMSE=\sqrt{\frac{1}{n}\sum_{i=1}^{n}\left(y_{pred_i}-y_{actual_i}\right)^2}
\label{rmse}
\end{equation}
Where $y_{pred_i}$ is the prediction value for the $i$-th sample, and the $y_{actual_i}$ is the actual value for the $i$-th sample. 
PyTorch was utilized, and the implementation of the prediction models referenced the code provided in \cite{LibCity}. A uniform configuration for hyperparameters, such as node embedding dimension and hidden layer dimension, was maintained across all training scenarios to ensure comparability of these calibrated models across different training scenarios. Our training strategy involved utilizing the Adam optimizer with an initial learning rate of 0.005. Each model underwent training for 100 epochs. Following each training epoch, the prediction error on the validation set was monitored. If the validation error did not decrease for five consecutive epochs, the learning rate was halved. Training was terminated prematurely if the validation error did not decrease for ten consecutive epochs. 
\subsection{Pre-experiment results}
In the Table \ref{pre result for bike} and Table \ref{pre result for taxi}, the test errors obtained on the NYCBIKE and NYCTAXI datasets in different training scenarios are reported (The blue cells indicate the minimum error in different training scenarios for each model)
\begin{table*}[!h]
\centering
\begin{tabular}{ccccccc}
\toprule[1.5pt]
Training month      & Metric & STGCN           & MTGNN           & AGCRN           & DCRNN           & GWNET           \\ \hline
\multirow{2}{*}{12} & MAE    & 6.456           & 6.555           & 6.530           & 6.433           & 6.597           \\
                    & RMSE   & 11.848          & 12.697          & 12.126          & 11.896          & 12.477          \\
\multirow{2}{*}{24} & MAE    & 6.251           & 6.321           & 6.372           & 6.393           & 6.571           \\
                    & RMSE   & 11.393          & 11.755          & 11.747          & 11.964          & 12.178          \\
\multirow{2}{*}{48} & MAE    & \cellcolor[HTML]{CBCEFB}\textbf{6.231}  & 6.245           & \cellcolor[HTML]{CBCEFB}\textbf{6.322}  & \cellcolor[HTML]{CBCEFB}\textbf{6.328}  & \cellcolor[HTML]{CBCEFB}\textbf{6.414}  \\
                    & RMSE   & 11.320 & 11.737          & 11.636 & 11.543 & 11.791 \\
\multirow{2}{*}{96} & MAE    & 6.267           & \cellcolor[HTML]{CBCEFB}\textbf{6.240}  & 6.361           & 6.342           & 6.428           \\
                    & RMSE   & 11.389          & 11.532 & 11.678          & 11.673          & 11.851   \\ \bottomrule      
\end{tabular}
\caption{Pre-experiment result of NYCBIKE dataset}
\label{pre result for bike}
\end{table*}

\begin{table*}[!h]
   \centering
\begin{tabular}{ccccccc}
\toprule
Training month      & Metric & STGCN          & MTGNN          & AGCRN          & DCRNN          & GWNET          \\ \hline
\multirow{2}{*}{12} & MAE    & 7.289          & 7.264          & 7.386          & 7.454          & 7.368          \\
                    & RMSE   & 13.247         & 13.106         & 13.420         & 13.373         & 13.293         \\
\multirow{2}{*}{24} & MAE    & 7.297          & \cellcolor[HTML]{CBCEFB}\textbf{7.264} & 7.252          & \cellcolor[HTML]{CBCEFB}\textbf{7.454} & \cellcolor[HTML]{CBCEFB}\textbf{7.452} \\
                    & RMSE   & 13.258         & 13.185         & 13.142         & 13.356         & 13.451         \\
\multirow{2}{*}{48} & MAE    & \cellcolor[HTML]{CBCEFB}\textbf{7.281}  & 7.267          &\cellcolor[HTML]{CBCEFB} \textbf{7.223} & 7.455          & 7.453          \\
                    & RMSE   & 13.209         & 13.125         & 13.065         & 13.496         & 13.507         \\
\multirow{2}{*}{96} & MAE    & 7.384          & 7.341          & 7.246          & 7.469          & 7.612          \\
                    & RMSE   & 13.405         & 13.319         & 13.137         & 13.544         & 13.787        \\ \bottomrule
\end{tabular}
\caption{Pre-experiment result of NYCTAXI dataset}
  \label{pre result for taxi}
\end{table*}
For the NYCBIKE dataset, the models generally show improved performance with increasing training data, but the results for the 96-month scenario do not consistently yield the best outcomes. STGCN achieves its lowest MAE and RMSE at 48 months (6.231 and 11.320, respectively). However, at 96 months, the MAE increases slightly to 6.267, and the RMSE also worsens to 11.389. MTGNN demonstrates its best performance at 96 months, but the improvements are minimal compared to 48 months. The MAE at 48 and 96 months remains close (6.245 vs. 6.240), with a slight RMSE improvement (11.737 vs. 11.532). AGCRN shows its lowest MAE and RMSE at 48 months, and its performance slightly degrades at 96 months (MAE of 6.361 and RMSE of 11.678). DCRNN does not benefit from the longer training period; its best performance is also at 48 months, with MAE increasing slightly at 96 months (6.342 vs. 6.328). GWNET performs worst among all models and the prediction error of it also shows the similar change pattern as other models when training with larger datasets.

For the NYCTAXI dataset, a similar trend can be observed: STGCN and MTGNN perform best at 48 months, with slightly increased MAE and RMSE at 96 months. AGCRN achieves the lowest errors at 48 months (MAE: 7.223, RMSE: 13.065), and both metrics degrade at 96 months (MAE: 7.246, RMSE: 13.137). DCRNN and GWNET also show worse performance at 96 months, highlighting that the additional training data does not necessarily lead to better results. Besides, the deterioration of prediction accuracy in NYCTAXI dataset is more serious than NYCBIKE dataset when training with larger datasets. Moreover, training with data from 96 months even leads to the worst prediction result for STGCN, MTGNN, DCRNN and GWNET. 

In summary, the analysis suggests that increasing the training dataset up to 96 months does not guarantee the best performance across all models. The trend highlights a contradictory effect, where larger datasets may lead to adverse returns in model improvement. Therefore, it can be concluded that devising some modules to make better use of long-term history data is necessary.
\section{Improved Training Method}
\subsection{Distribution Shift Analysis}
\begin{figure*}
    \centering
    \includegraphics[width=\linewidth]{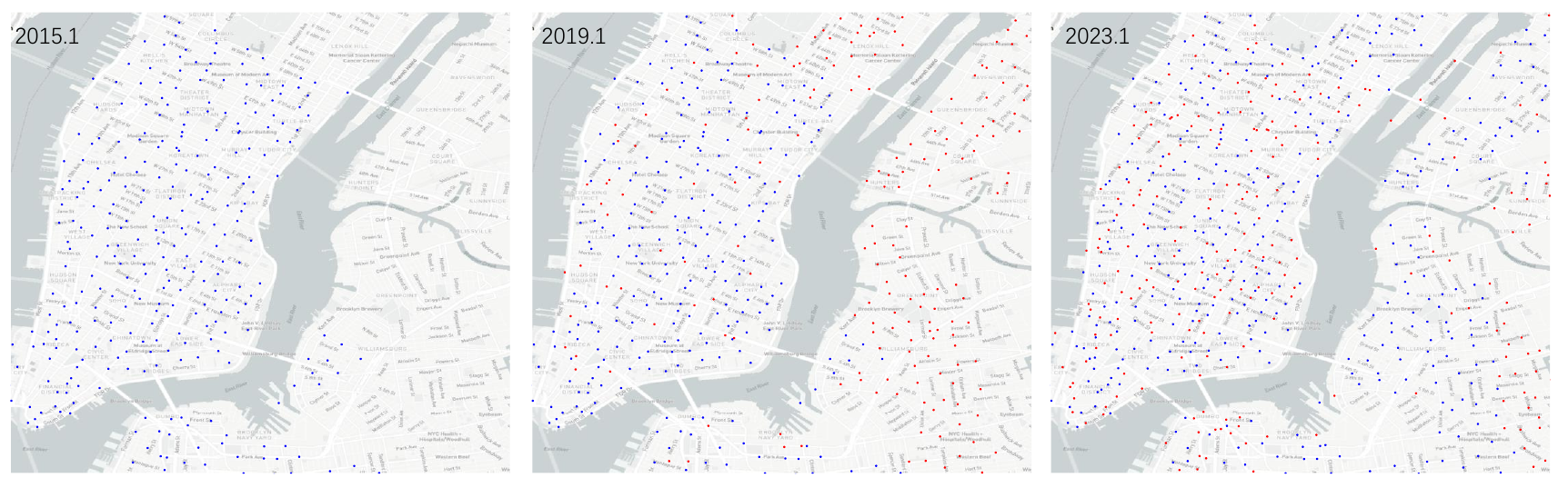}
    \caption{Bike station distribution in different years (Blue points means existing bike stations, red points means the stations newly opened in the last 4 years)}
    \label{fig:Bike station distribution}
\end{figure*}

The phenomenon that more data sometimes causes more error might be attributed to disparities between patterns observed in earlier historical data and those present in the test set. For example, New York's shared bikes are station-based, and Figure \ref{fig:Bike station distribution} illustrates the distributions of bike stations in Manhattan in January 2015, January 2019, and January 2023. It is evident that the number of bike stations has increased over time, leading to changes in bike usage patterns. Similarly, for taxi usage, the rise of online ride-hailing services over the eight-year period could contribute to changes in usage patterns. Besides, the COVID-19 could also affect bike and taxi usage patterns. Consequently, patterns learned from data in prior years may not be directly applicable to the test set. 

Before further analysis, it is necessary to revisit the rationale behind the traditional machine learning that minimizing training set errors can obtain a model with comparatively lower errors on the test set. The reasons for adopting this approach will be elucidated in the following.

Initially, we denote that the data distribution of the test set as $P_{te (x,y)}$, while the distribution in the training dataset is denoted as $P_{tr (x,y)}$. To ensure that our model performs well on the test dataset, $E_{(x,y) \sim P_{te(x,y)}} (loss(f_\theta(x),y))$ need to be minimized, where the $f_\theta (x)$ represents a predictive model with parameters $\theta$ and input $x$. In the conventional machine learning scenario, the data distributions of the training and test sets are assumed to be the same, implying $P_{tr(x,y)}=P_{te (x,y)}$. Therefore, the goal can be changed as:

\begin{equation}
\mathop{min}_{\theta}E_{(x,y)\sim P_{tr(x,y)}}(loss(f_\theta ((x),y))
\end{equation}
This expectation is approximated using samples from the training dataset, leading to the ultimate training objective: 
\begin{equation}
    \sum_{(x,y)\in D_{train}}loss(f_\theta ((x),y)
\end{equation}
Where $D_{train}$ is the training dataset. As a result, by minimizing the loss on the training set, a model with smaller losses on the test set can usually\footnote{We use “usually” because there is generalization gap between training error and testing error which may cause the models with smaller training error to perform worse in test set, but this is beyond our focus.} be obtained.

While in our scenario, due to potential changes over time, the sample distributions in the training and testing sets may differ, indicating $P_{tr (x,y)}\neq P_{te (x,y)}$. Therefore, employing a straightforward optimization of the loss in training set cannot result in a model with smaller error on the test set. The method addressing the issues related to different sample distributions between the training and testing sets is known as transfer learning. In some transfer learning literature \cite{datasetshift}, the joint distribution of $x$ and $y$ can be decomposed into two components: the marginal distribution of x and the conditional distribution of $y$ given $x$, which can be expressed as: $P(x,y) = P(x)P(y|x)$. When $P_{tr (x)}\neq P_{te (x)}$, this situation is named as covariate shift. If $P_{tr (x|y)}\neq P_{te (x|y)}$ occurs, it is termed as concept shift\footnote{Concept shift is called concept drift in some papers \cite{Li2022DDGDADD,zhang2023onenet}.}. And different modules can be deployed to address different types of distribution shift. For example, in time series prediction task, \cite{Du2021AdaRNNAL,Kim2022ReversibleIN} aims to address covariate shift and \cite{Li2022DDGDADD,zhang2023onenet} commit to address concept shift. For our problem, both types of shifts are present.
\begin{figure*}[!h]
\centering
    \includegraphics[width=\linewidth]{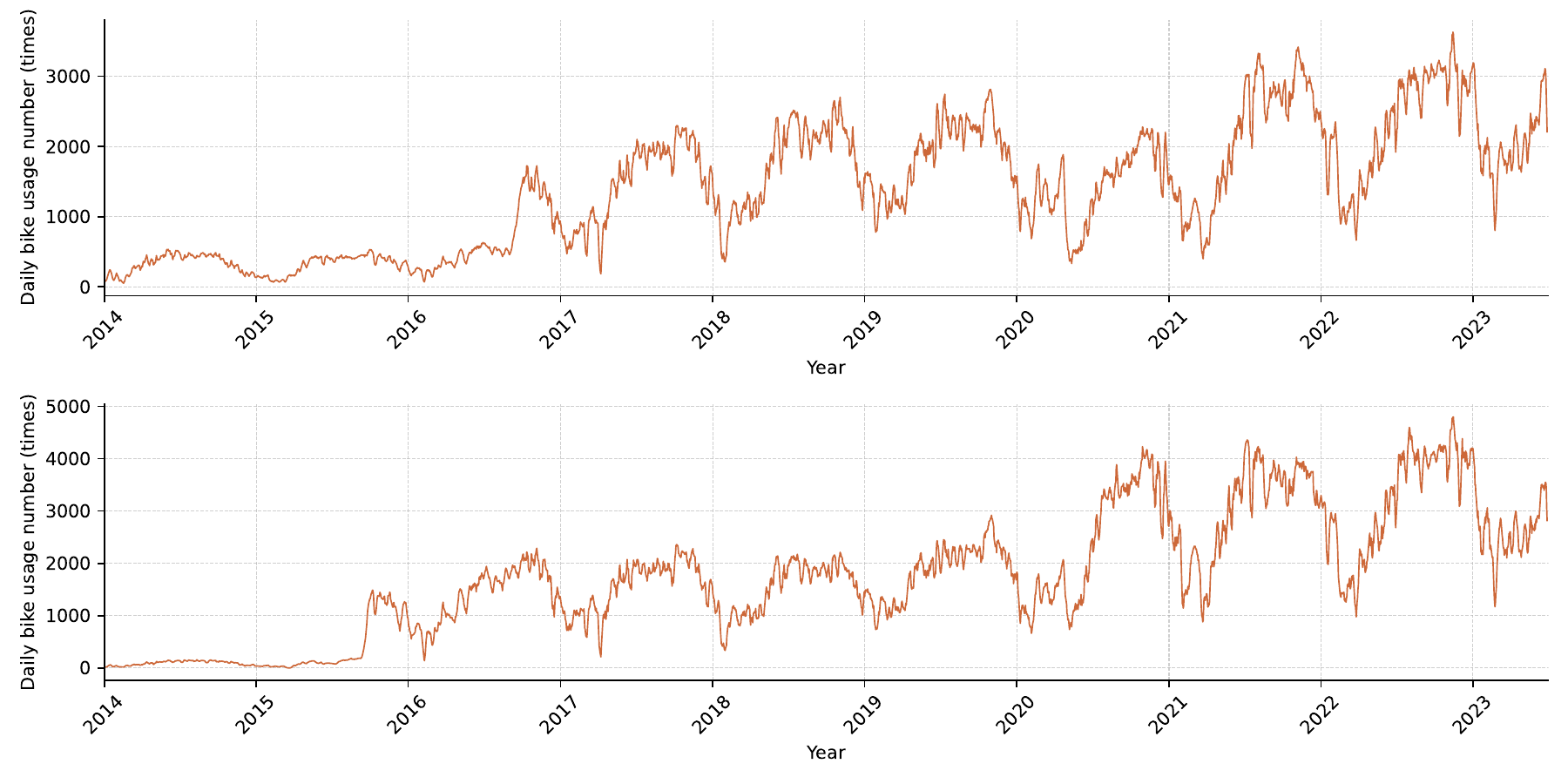}
    \caption{The change of bike usage numbers in two different girds}
    \label{fig:bike_usage}
\end{figure*}
\begin{figure*}[!h]
\centering
    \includegraphics[width=\linewidth]{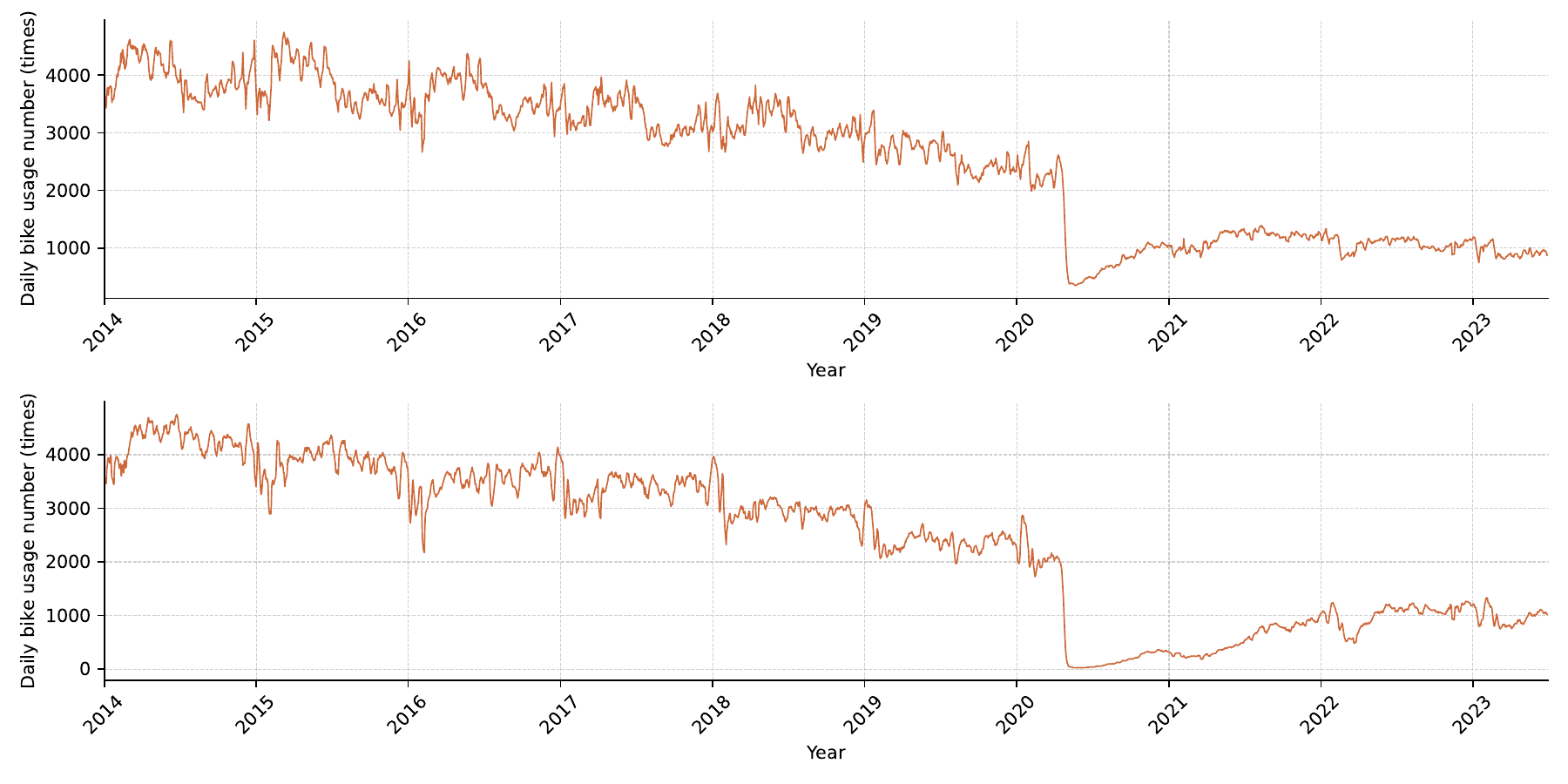}
    \caption{The change of taxi usage numbers in two different regions}
    \label{fig:taxi_usage}
\end{figure*}

To illustrate covariate shift and concept shift, we will provide examples for both cases. Regarding the covariate shift, there has been a continuous increase in the number of bike stations, resulting in an overall upward trend in bike usage in each grid. For example, Figure \ref{fig:bike_usage} plots the daily usage of shared bikes for two grids from 2014 to 2022, revealing an increasing trend in bike usage. Given that our model incorporates bike usage from the preceding time steps as input, it becomes apparent that in training set the input covariates of the model are relatively smaller, whereas in the test set, the input covariates tend to be larger. Besides, the daily taxi usage number of some regions are also plotted in Figure \ref{fig:taxi_usage}, a significant drop can be found at around 2020 and that might be highly relate to the COVID-19, which also implies the covariate shift.

Regarding concept shift, let's envision a scenario where a new entertainment facility opens in a specific area. This facility may not significantly impact daytime traffic volume on weekdays but could lead to a substantial increase in traffic volume during the night. Consequently, as our model considers traffic volume from the previous six hours as input and predicts the traffic volume for the next hour, if the former six hours correspond to daytime and the next hour to nighttime, then even though the traffic volume for the preceding six hours remains the same, the traffic volume in the later hour might be different. This phenomenon exemplifies concept shift.

Therefore, we have devised a unique method that simultaneously addresses both covariate shift and concept shift. Besides, it is also adaptable across most existing short-term traffic prediction models. To address concept shift, we carefully analysis the cause of concept shift. The inconsistency between long-term historical data and present data may be rooted in changes of background variables like land usage and economic conditions. Identifying and incorporating these variables into the model might be able to transfer the out-of-distribution history data into in-distribution data. Mathematically, noting the data as $(x,y)$ and considering$ P(y|x)_{history}\neq P(y|x)_{now}$. However, with the background knowledge $z$, the distribution conditional on $z$ might be the same, which can be expressed as Eq.\ref{eq:z}:
\begin{equation}
    P(y|x,z)_{history}=P(y|x,z)_{now}
    \label{eq:z}
\end{equation}

Therefore, we draw inspiration from environment aware learning \cite{Environment-Aware}, and try to learn the background knowledge in different period from data. Moreover, in order to address covariate shift, our approach employs a straightforward weighting scheme applied to historical data, aligning the distribution of covariate in historical data with that in future data.
\subsection{Module for Concept Shift}
To mitigate the impact of concept shift, inspiration is drawn from environment aware learning. This method means the data is from different environments (i.e., time periods) and we need to learn the features (such as land usage information) in different environments and incorporate them to the model. However, as me mentioned before, the traffic prediction task is of some distinct features and we propose methods to address them in the following parts.

\subsubsection{Environment feature learning for each region}

One distinct feature in traffic prediction problem is that the environment feature changes in different regions are different. For example, if some facilities are opened in one region, then then environment feature for this region will change but the environment features for other regions might hold unchanged. Therefore, we need to address the environment change specifically for each region. One method is collecting the features such as land usage and population information for each region and each time period and directly incorporate them into model. However, it is difficult to get access to that information. Another solution is to use changing-point detection algorithm for the traffic volume in each region and obtain the changing points for each region, then use some method to tackle these changes. The drawback of these method is that it could result in a two-stage algorithm, so the efficiency could be impaired. Besides, the accuracy of changing point detection could influence the final result. Therefore, we design a more flexible method to concurrently learn the changing points and environment features for each region. We will elaborate in follows.

We set each month as an environment and establish an environment feature pool for all regions, for example, if there are $n$ regions, $m$ months and each feature for an environment and a region is of dimension $d$, then we will have an environment feature pool of shape $n\times m\times d$. And we set these parameters learnable and incorporate these features in traffic prediction models. Therefore, these parameters can be learned through model training process. However, this intuitive method could lead to too many learnable parameters and we choose to impose some prior information to restrict the freedom these parameters.

\begin{figure}[!h]
    \centering
    \includegraphics[width=0.9\linewidth]{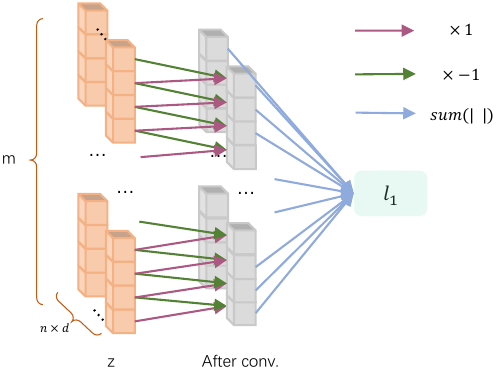}
    \caption{The calculation of regular term}
    \label{fig:regular}
\end{figure}

It is known that the environment of a region does not change so frequently, which means that the environment features for one region are basically same for two consecutive months. Only of some cases, these environment features changed (such as new supermarket opens). Therefore, a regulation term is devised to consider this pattern, as shown in Figure \ref{fig:regular}. We reshape environment feature pool as a matrix $Z$ of shape $( n\times d)\times m$, and use the following Eq.\ref{regular} to obtain this term.
\begin{equation}
    l_1=\sum |(1,-1)*Z|
    \label{regular}
\end{equation}
\subsubsection{Region relationship learning in different environments}
As analyzed before, the relationship between regions might change through time because of the change of land usage or some other factors. To verify this statement further, we use the data about bike and taxi usage from 2015 to 2016, from 2017 to 2018, from 2019 to 2020 and from 2021 to 2022 to train different AGCRN models. The AGCRN model is able to learn adaptive adjacent matrix from data. Therefore, we can compare the learned adjacent matrices from these different models. These matrices are plotted in the following Figure \ref{fig:taxi_adj} and Figure \ref{fig:taxi_adj}. The first figure is the adjacent matrix for bike dataset, and the second is for taxi dataset.
\begin{figure}
    \centering
    \includegraphics[width=0.95\linewidth]{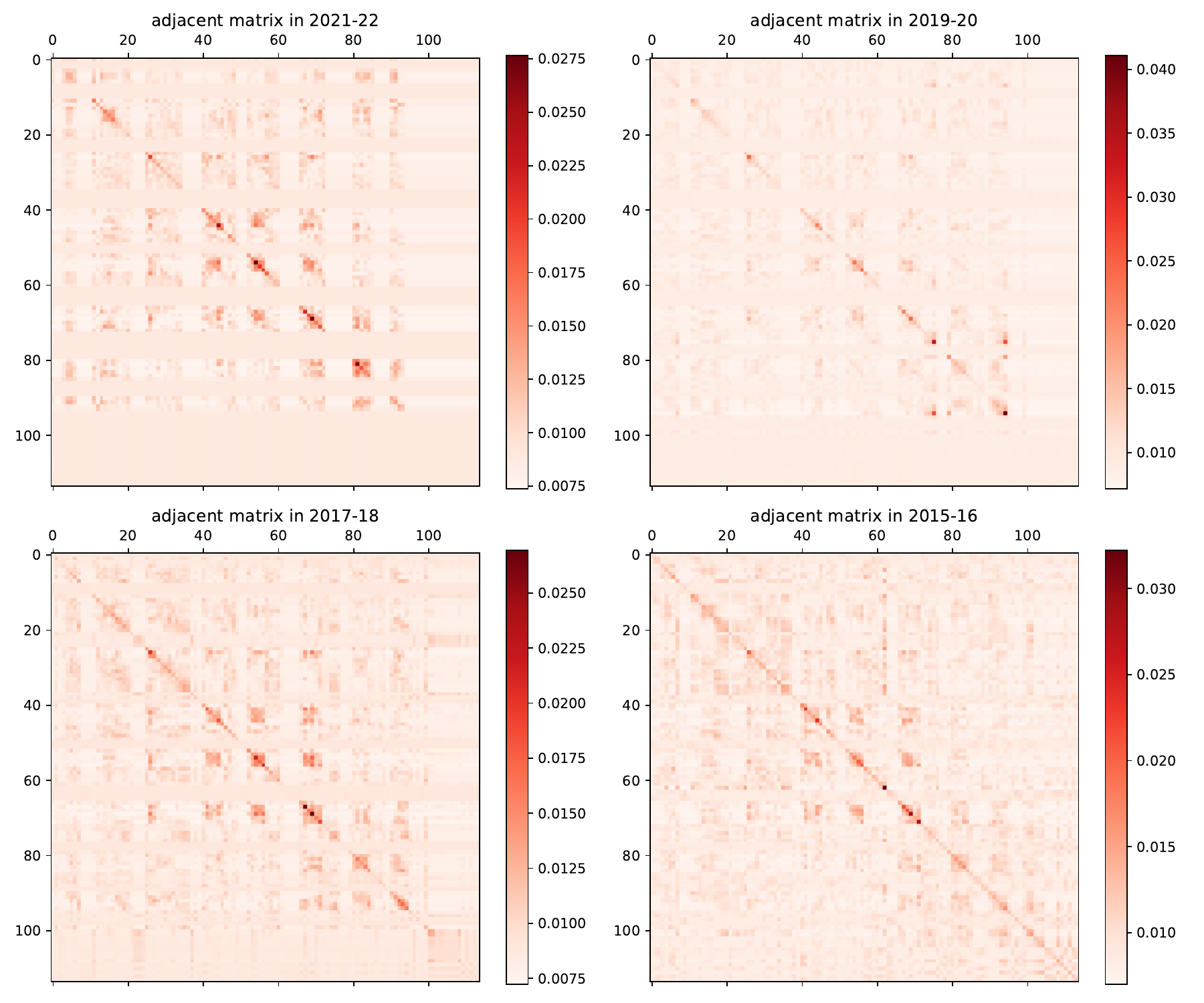}
    \caption{The adjacent matrices learned in different time periods for NYCBIKE dataset}
    \label{fig:bike_adj}
\end{figure}
\begin{figure}
    \centering
    \includegraphics[width=0.95\linewidth]{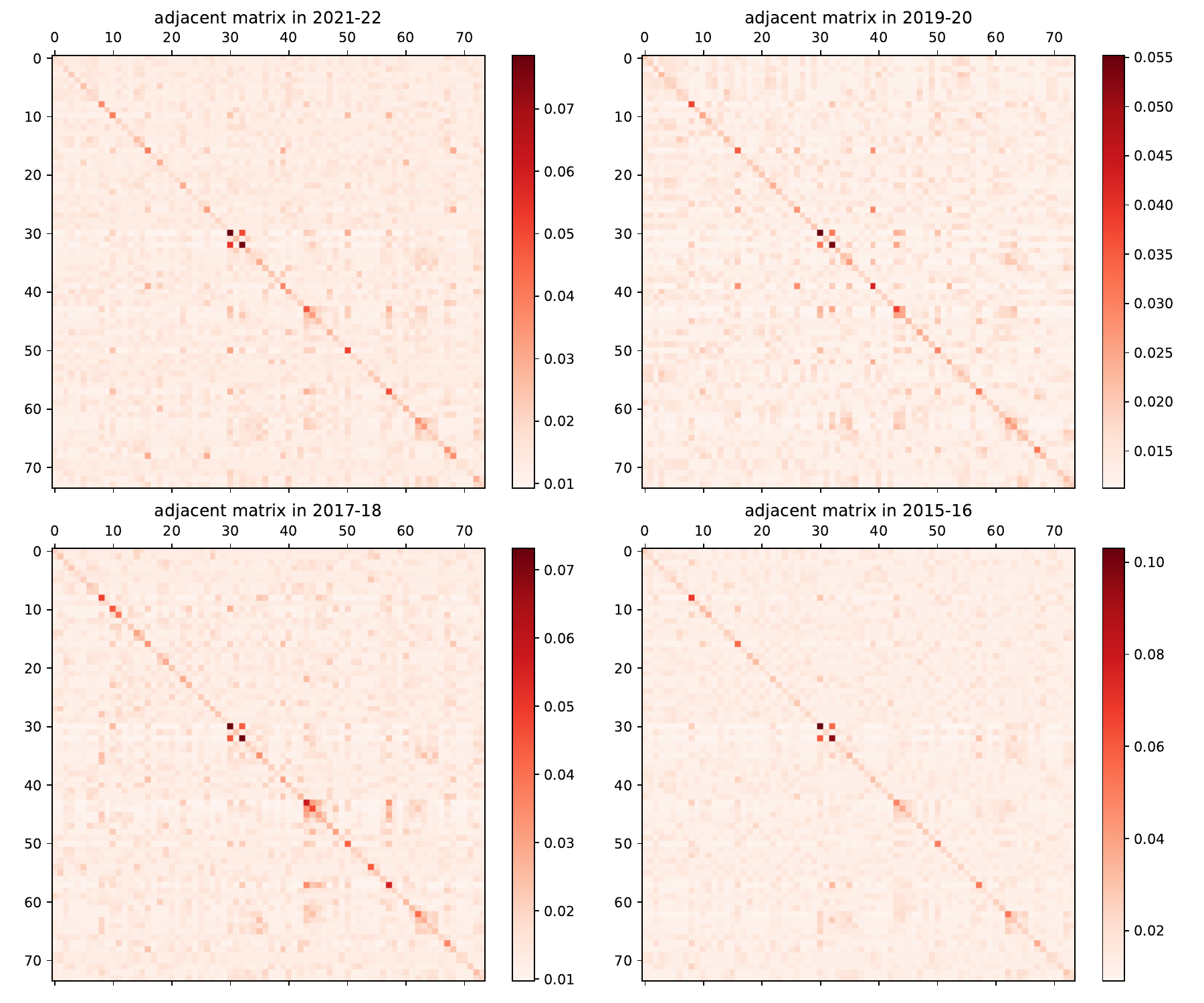}
    \caption{The adjacent matrices learned in different time periods for NYCTAXI dataset}
    \label{fig:taxi_adj}
\end{figure}

It can be observed from the Figure \ref{fig:bike_adj} and Figure \ref{fig:taxi_adj} that there are different adjacent matrices in different period. Therefore, we can image a situation that there were correlations between two regions i and j during 2015 to 2020, but the relationship vanished in 2021. And if a model is trained with data from 2015 to 2022, then it will be likely for the model to learn that region i and j should be connected. However, this learned adjacent matrix reflects incorrect spatial relationship after 2021. Then when the model is deployed to predict traffic volumes in 2022, the accuracy will be deteriorated because of the improper adjacent matrix. As a result, devising method to tackle this issue is necessary.

Therefore, we propose a method to fit different adjacent matrix in different period. In detail, we create an embedding vector for every region in each month, then the vectors for all regions in one month can be concatenated as a matrix $M\in \mathbb{R}^{n\times d_2}$, where n is the number of regions and $d_2$ is the dimension of embedding vector. The adjacent can be obtained by Eq.\ref{eq:adj} subsequently:
\begin{equation}
    A=softmax(M^TM)
    \label{eq:adj}
\end{equation}
The soft-max function is used to ensure that each element in A is in (0,1). Besides, it should be noticed that the adjacent matrix could not change substantially in a short period, therefore we design a regular term to ensure that adjacent matrices in short time difference can not differ significantly. This regular term is similar to Eq.\ref{regular}. Specifically, all M's are stacked together and reshape it to a matrix with shape $(n\times m)\times d_2$ and we also use Eq.\ref{regular} to obtain the regular term, which is donated as $l_2$.

In summary, given the node feature and adjacent matrix in each month as environment features z, the distribution output y condition on z and input x can be considered the same in all periods. The remaining problem is how to incorporate these features into existing traffic prediction models and we will elaborate it in the model structure section.
\subsection{Module for Covariate Shift}
The definition of covariate shift entails different distributions of independent variables between the training and test datasets. Therefore, the most intuitive approach to address covariate shift is to weight the samples, assigning higher weights to the training samples which are more likely to appear in the test set. This method ensures that the distribution of independent variables in the weighted training set is similar to the test set [26]. Consequently, a classifier is trained to output the probability for each training sample to appear in test set. And the classifier itself can be constructed by transforming a spatiotemporal neural network used for prediction into one for classification (e.g., by adding a binary linear layer to the last layer of the neural network)

However, the test set cannot be acquired when training. To solve this problem, we regard the data from the most recent 12 month in training set as a proxy for test set because it is reasonable to assume that the data from the most recent months is much similar to the test set. Therefore, positive samples are derived from data in the most recent 12 months, while the older data is considered as negative samples. Then a classifier can be trained using these datasets. Therefore, the classifier can output the probability for each sample to appear in test set.

We should consider two aspects when weighting training samples in the time series prediction task: one is similarity \cite{Lu2021RethinkingIW}, the data more similar to test data need to be with larger weights, the other is recency, the data in more recent month need to be with larger weights. Therefore, if the probability for a sample $(x_i,y_i )$ to appear in test set is predicted as $p_i$ according to the classifier, we use the following Eq.\ref{eq:w} to give the sample a weight.
\begin{equation}
    w_i=p_i (1-(\delta_i-1)/\Delta)^\beta
    \label{eq:w}
\end{equation}

Where $w_i$ is the weight for $(x_i,y_i )$, $\delta_i$ is the time difference (countered by month number) between the sample and test month. For example, if test month is 2023.6 and the training data is from 2022.6, then $\delta_i$  will equal 12. $\Delta $is the number of months included in training set. The term can $1-(\delta_i-1)/\Delta$ can ensure the sample from earlier time tends to be given with smaller weight. And $\beta$ is used to control the balance of similarity and recency.
\subsection{Model Structure, Training and Deploying}
We hope that our method can be adapted to most existing spatial temporal models. As a result, we do not choose to change the structure of existing models but change the date feeding process.
\begin{figure*}[!h]
    \centering
    \includegraphics[width=0.95\linewidth]{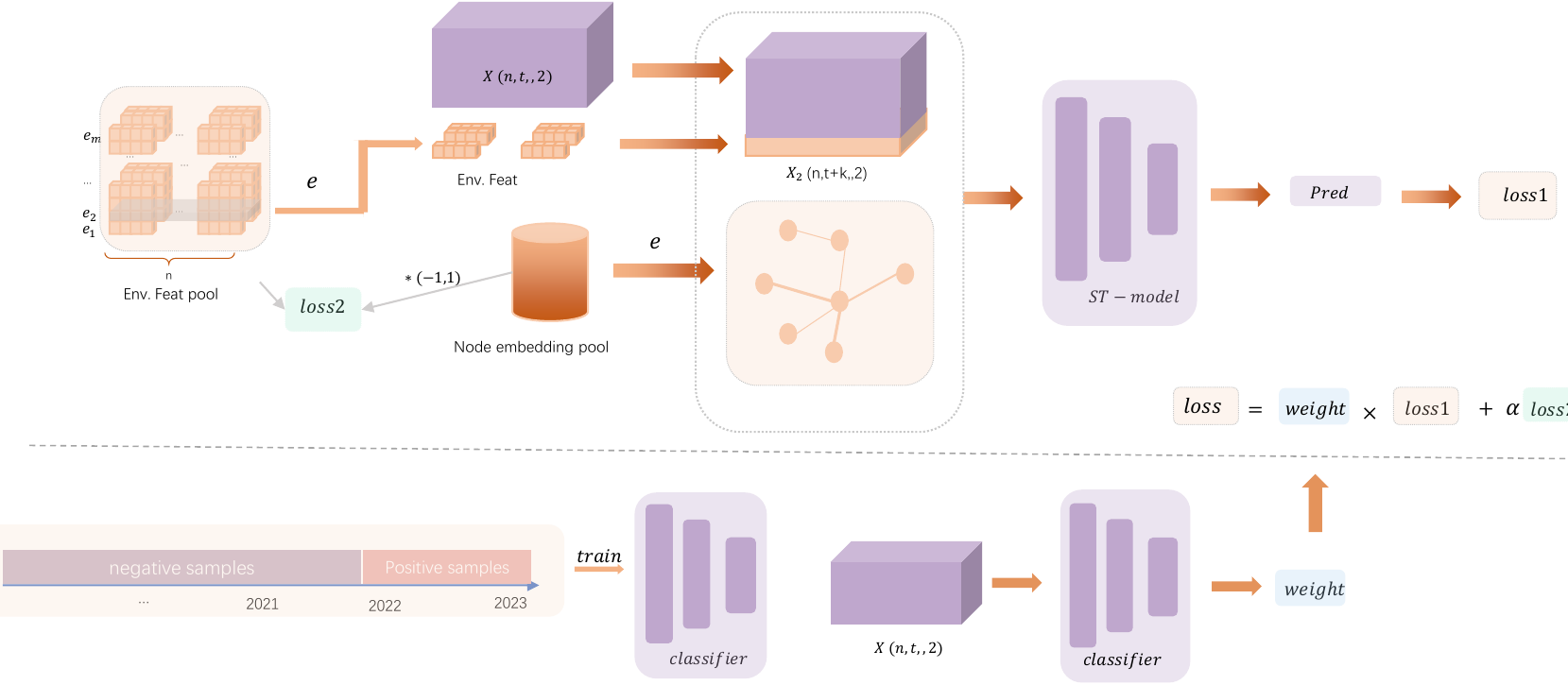}
    \caption{The overview of our method}
    \label{fig:workflow}
\end{figure*}
For an original spatial-temporal model, the history data X and the adjacent matrix A need to be fed into it and the prediction Y can be obtained, then the loss can be calculated. In our method, as shown in Figure \ref{fig:workflow}, if the input history data $X$ is of shape $(n,t,2)$ and from the environment $e$, then we can use $e$ to retrieve the environment feature (Env.Feat in Figure \ref{fig:workflow}) from the environment feature pool (Env.Feat Pool in Figure \ref{fig:workflow}). Then the feature is reshaped to $(n,d/2,2)$ and concatenated to $X$, then the input $X_2$ is obtained of the shape $(n,t+d/2,2)$. In another word, this method equals to add addition time slices to $X$, i.e., reshaping the second dimension of $X$ from $t$ to $t+d/2$. For the models using RNN, such as DCRNN and AGCRN, this method equals to use a specific initialization of hidden state. Because for traditional RNN, the hidden state is initialized with a constant vector, but we add some additional time slices, so after the RNN model goes through this additional time slices, the hidden states may be changed and this change is related to the data in additional time slices, in another word, the data in environment features.

Besides, using the environment $e$, the embeddings from the node embedding pool can also be retrieved, and these node embeddings can be used to generate an adjacent matrix $A$. Afterwards, the input tensor $X_2$ obtained from the formal step and the adjacent matrix $A$ can be fed into the traffic prediction model together, and the final prediction result can be obtained.

Moreover, before training the traffic prediction model, a classification model is trained with samples from the recent one year as positive samples and samples from earlier times as negative samples. Then the prediction results of the classification model can be used to give weight to each sample according to Eq.\ref{eq:loss}. Finally, the loss function can be written as:
\begin{equation}
    loss=|y_{pred}-y_{true} |\times w+\alpha(l_1+l_2)
    \label{eq:loss}
\end{equation}
Where $w$ is the sample weight, $y_{pred}$ and $y_{true}$ are the prediction and the ground truth, $l_1$ and $l_2$ are regularization terms as we define in the formal section and the summation of them is expressed as $loss_2$ in Figure 
\ref{fig:workflow}.

When deploying, we use the environment features of the last month in training dataset as the environment feature of the test set. Also, the learnt adjacent matrix in the last month in training set is used when testing.
\section{Experiments}
Some details in our experiment are first described as follows. The dimension of environment features and node embeddings are both set at 4. We use grid search to find the best $\alpha$ and $\beta$, the search spaces are both 0.1,1,10. Besides, only predefined adjacent matrix based on geographic adjacency is used in STGCN and DCRNN. Therefore, we do not integrate dynamic learnable adjacent matrix module in these two models.

The dataset, models and metrics are the same as pre-experiment section, so we omit it in this section.
\subsection{Baseline}
It should be noticed that there is no method designed specifically to train traffic prediction model with long term historical data. Besides, in the domain of general time series prediction, there are some methods proposed to solve the distribution shift problem. Therefore, we select some of these methods as our baselines.
\begin{enumerate}
    \item \textbf{AdaRNN} \cite{Du2021AdaRNNAL}: The main idea for this method is regarding outputs of hidden layers in RNN as features of covariate and use MMD loss \cite{Pan2009DomainAV} to align these features. As a result, the feature distribution of data in different time periods can be aligned and the issue of covariate shift can be addressed. In our experiments, DCRNN, AGCRN contains RNN layers and can be directly used in AdaRNN method. However, there are on RNN layers in GWNET, STGCN and MTGNN, therefore, we reconstruct the AdaRNN method and regard the output of the first two spatial temporal blocks as the features of covariates and use MMD loss to align these features.
    \item \textbf{RevIN} \cite{Kim2022ReversibleIN}: This method aligns the distributions of covariate in different periods by shifting and scaling the data to make the mean and variance of training data in different time periods the same.
    \item \textbf{IRM} \cite{Arjovsky2019InvariantRM}: This method assumes the dataset is from different domains but there are common causality relationships in different domains and attempts to keep the risk in different domains invariant to force the model to learn the common causality relationships. We use the variances of loss in different periods as a regular term to realize this target according to \cite{pmlr-v139-krueger21a}.
    \item \textbf{Finetune}: Finetuning is commonly used in transfer learning. When finetuning, the data from the most recent 12 months are used to finetune the model for 20 epochs with learning rate set as $10^{-4}$.
\end{enumerate}
\subsection{Result}
The results of our method and the baselines are described in the following Table \ref{table:result1} and Table \ref{table:result2}.
\begin{table*}[!h]
\centering
\begin{tabular}{@{}cccccccccc@{}}
\toprule
\multirow{2}{*}{Training month} & \multirow{2}{*}{Model} & \multirow{2}{*}{Metric} & \multirow{2}{*}{Original} & \multirow{2}{*}{AdaRNN} & \multirow{2}{*}{IRM} & \multirow{2}{*}{Finetune} & \multirow{2}{*}{RevIN} & \multirow{2}{*}{Our method} & \multirow{2}{*}{Improvement\footnote{The improvements in Table \ref{table:result1} and Table \ref{table:result2} are calculated based on MAE}} \\ 
                                &                        &                                  &                           &                         &                      &                           &                        &                             &                              \\ \midrule
\multirow{10}{*}{48}            & \multirow{2}{*}{STGCN} & MAE                              & 6.231                     & 6.21                    & 6.249                & 6.174                     & 6.273                  & \textbf{6.094}              & \multirow{2}{*}{2.2\%}       \\
                                &                        & RMSE                             & 11.32                     & 11.293                  & 11.529               & 11.24                     & 11.502                 & \textbf{11.137}             &                              \\
                                & \multirow{2}{*}{MTGNN} & MAE                              & 6.245                     & 6.231                   & 6.244                & 6.22                      & 6.3                    & \textbf{6.119}              & \multirow{2}{*}{2.0\%}       \\
                                &                        & RMSE                             & 11.737                    & 11.522                  & 11.523               & 11.438                    & 11.526                 & \textbf{11.219}             &                              \\
                                & \multirow{2}{*}{AGCRN} & MAE                              & 6.322                     & 6.31                    & 6.315                & 6.289                     & 6.342                  & \textbf{6.173}              & \multirow{2}{*}{2.4\%}       \\
                                &                        & RMSE                             & 11.636                    & 11.482                  & 11.485               & 11.42                     & 11.602                 & \textbf{11.462}             &                              \\
                                & \multirow{2}{*}{DCRNN} & MAE                              & 6.328                     & 6.334                   & 6.36                 & 6.313                     & 6.318                  & \textbf{6.255}              & \multirow{2}{*}{1.2\%}       \\
                                &                        & RMSE                             & 11.543                    & 11.642                  & 11.645               & 11.683                    & 11.521                 & \textbf{11.354}             &                              \\
                                & \multirow{2}{*}{GWNET} & MAE                              & 6.414                     & 6.377                   & 6.38                 & 6.371                     & 6.425                  & \textbf{6.265}              & \multirow{2}{*}{2.3\%}       \\
                                &                        & RMSE                             & 11.791                    & 12.013                  & 12.013               & 11.744                    & 12.037                 & \textbf{11.465}             &                              \\ \midrule
\multirow{10}{*}{96}            & \multirow{2}{*}{STGCN} & MAE                              & 6.267                     & 6.224                   & 6.215                & 6.189                     & 6.218                  & \textbf{6.062}              & \multirow{2}{*}{3.3\%}       \\
                                &                        & RMSE                             & 11.389                    & 11.33                   & 11.402               & 11.301                    & 11.376                 & \textbf{11.048}             &                              \\
                                & \multirow{2}{*}{MTGNN} & MAE                              & 6.24                      & 6.221                   & 6.257                & 6.263                     & 6.346                  & \textbf{6.006}              & \multirow{2}{*}{3.8\%}       \\
                                &                        & RMSE                             & 11.532                    & 11.451                  & 11.59                & 11.508                    & 11.744                 & \textbf{11.216}             &                              \\
                                & \multirow{2}{*}{AGCRN} & MAE                              & 6.361                     & 6.292                   & 6.303                & 6.29                      & 6.384                  & \textbf{6.153}              & \multirow{2}{*}{3.3\%}       \\
                                &                        & RMSE                             & 11.678                    & 11.726                  & 11.93                & 11.674                    & 11.838                 & \textbf{11.201}             &                              \\
                                & \multirow{2}{*}{DCRNN} & MAE                              & 6.342                     & 6.332                   & 6.329                & 6.314                     & 6.352                  & \textbf{6.148}              & \multirow{2}{*}{3.1\%}       \\
                                &                        & RMSE                             & 11.673                    & 11.65                   & 11.493               & 11.657                    & 11.622                 & \textbf{11.419}             &                              \\
                                & \multirow{2}{*}{GWNET} & MAE                              & 6.428                     & 6.376                   & 6.409                & 6.357                     & 6.447                  & \textbf{6.22}               & \multirow{2}{*}{3.2\%}       \\
                                &                        & RMSE                             & 11.851                    & 11.993                  & 11.55                & 11.724                    & 12.073                 & \textbf{11.618}             &                              \\ \bottomrule
\end{tabular}
\caption{Experiment result for NYCBIKE dataset}
\label{table:result1}
\end{table*}

\begin{table*}[!h]
\centering
\begin{tabular}{@{}cccccccccc@{}}
\toprule
\multirow{2}{*}{Training month} & \multirow{2}{*}{Model} & \multirow{2}{*}{Metric} & \multirow{2}{*}{Original} & \multirow{2}{*}{AdaRNN} & \multirow{2}{*}{IRM} & \multirow{2}{*}{Finetune} & \multirow{2}{*}{RevIN} & \multirow{2}{*}{Our method} & \multirow{2}{*}{Improvement} \\
                                &                        &                         &                           &                         &                      &                           &                        &                             &                              \\ \midrule
\multirow{10}{*}{48}            & \multirow{2}{*}{STGCN} & MAE                     & 7.281                     & 7.239                   & 7.283                & 7.161                     & 7.186                  & \textbf{7.119}              & \multirow{2}{*}{2.2\%}       \\
                                &                        & RMSE                    & 13.209                    & 13.133                  & 13.258               & 13.009                    & 13.06                  & \textbf{12.861}             &                              \\
                                & \multirow{2}{*}{MTGNN} & MAE                     & 7.267                     & 7.227                   & 7.234                & 7.199                     & 7.181                  & \textbf{7.104}              & \multirow{2}{*}{2.2\%}       \\
                                &                        & RMSE                    & 13.125                    & 13.096                  & 13.161               & 12.994                    & 12.948                 & \textbf{12.832}             &                              \\
                                & \multirow{2}{*}{AGCRN} & MAE                     & 7.223                     & 7.204                   & 7.197                & 7.199                     & 7.166                  & \textbf{7.035}              & \multirow{2}{*}{2.6\%}       \\
                                &                        & RMSE                    & 13.065                    & 13.035                  & 13.052               & 13.068                    & 12.973                 & \textbf{12.734}             &                              \\
                                & \multirow{2}{*}{DCRNN} & MAE                     & 7.455                     & 7.341                   & 7.414                & 7.345                     & 7.336                  & \textbf{7.272}              & \multirow{2}{*}{2.5\%}       \\
                                &                        & RMSE                    & 13.496                    & 13.216                  & 13.374               & 13.26                     & 13.235                 & \textbf{13.051}             &                              \\
                                & \multirow{2}{*}{GWNET} & MAE                     & 7.453                     & 7.46                    & 7.416                & 7.342                     & 7.421                  & \textbf{7.231}              & \multirow{2}{*}{3.0\%}       \\
                                &                        & RMSE                    & 13.507                    & 13.742                  & 13.445               & 13.287                    & 13.482                 & \textbf{13.08}              &                              \\ \midrule
\multirow{10}{*}{96}            & \multirow{2}{*}{STGCN} & MAE                     & 7.384                     & 7.239                   & 7.393                & 7.312                     & 7.294                  & \textbf{7.072}              & \multirow{2}{*}{4.2\%}       \\
                                &                        & RMSE                    & 13.405                    & 13.133                  & 13.468               & 13.366                    & 13.222                 & \textbf{12.704}             &                              \\
                                & \multirow{2}{*}{MTGNN} & MAE                     & 7.341                     & 7.221                   & 7.321                & 7.201                     & 7.315                  & \textbf{7.079}              & \multirow{2}{*}{3.6\%}       \\
                                &                        & RMSE                    & 13.319                    & 13.278                  & 13.305               & 12.997                    & 13.278                 & \textbf{12.751}             &                              \\
                                & \multirow{2}{*}{AGCRN} & MAE                     & 7.246                     & 7.176                   & 7.244                & 7.199                     & 7.263                  & \textbf{7.005}              & \multirow{2}{*}{3.3\%}       \\
                                &                        & RMSE                    & 13.137                    & 13.073                  & 13.113               & 13.051                    & 13.09                  & \textbf{12.615}             &                              \\
                                & \multirow{2}{*}{DCRNN} & MAE                     & 7.469                     & 7.449                   & 7.418                & 7.411                     & 7.443                  & \textbf{7.27}               & \multirow{2}{*}{2.7\%}       \\
                                &                        & RMSE                    & 13.544                    & 13.562                  & 13.475               & 13.422                    & 13.328                 & \textbf{13.176}             &                              \\
                                & \multirow{2}{*}{GWNET} & MAE                     & 7.612                     & 7.588                   & 7.608                & 7.416                     & 7.695                  & \textbf{7.201}              & \multirow{2}{*}{5.4\%}       \\
                                &                        & RMSE                    & 13.787                    & 13.785                  & 13.685               & 13.445                    & 13.954                 & \textbf{13.234}             &                              \\ \bottomrule
\end{tabular}
\caption{Experiment result for NYCTAXI dataset}
\label{table:result2}
\end{table*}
For the NYCBIKE dataset, with a 48-month training period, our method shows significant improvements in both MAE and RMSE metrics for each model. For example, our method achieves a 2.2\% improvement in MAE compared to the original method (6.094 vs. 6.231) for STGCN model, and a notable reduction in RMSE (11.137 vs. 11.320) can also be observed. Similar trends are found across other models: MTGNN sees a 2.0\% improvement in MAE, and AGCRN and GWNET show improvements of 2.4

Moving to the 96-month training period, the improvements are even more pronounced. Our method consistently achieves the lowest errors compared to the baselines. For instance, in the STGCN model, the MAE drops by 3.3\% compared to the original method (6.062 vs. 6.267). Other models like MTGNN, DCRNN AGCRN and GWNET, the MAE improvements reach 3.8\% and 3.1\%. Besides, it can be found that the improvements for method using adaptive adjacent matrix (MTGNN, AGCRN and GWNET) are larger than the other models (STGCN and DCRNN).

For the NYCTAXI dataset, similar patterns emerge. In the 48-month setting, our method leads to improvements across all models, with the most notable being GWNET (3.0\% improvement in MAE) and AGCRN (2.6\% improvement in MAE). In the 96-month setting, the gaps between our method and the baselines widen further. For example, in the GWNET model, the MAE improves by 5.4\% compared to the original method (7.201 vs. 7.612), with a substantial RMSE reduction. Likewise, STGCN sees a 4.2\% improvement in MAE, and MTGNN achieves a 3.6\% improvement.

In summary, although the baseline methods, AdaRNN, IRM, RevIN and finetune, can decrease the prediction error, the improvements achieved by our method are more significant. This is because even though these baselines are designed to address distribution shift, but they are not proposed specific to do spatial-temporal prediction. As a result, they are not able to tackle distribution shift in the traffic prediction domain well. Besides, the improvements in 96 months cases are much more pronounced than 48 months. The reason might be that the distribution shift is more obvious in larger dataset, so the benefits of our method is more distinct. Finally, the errors in 96 months are all smaller than the errors in 48 months when using our method, which shows that our method can help models benefit from increased historical data.

\subsection{Ablation experiments}

To vindicate modules for concept shift and covariate shift are both important, we conduct the ablation experiments, which means we trained models without concept shift module and covariate shift module. Besides, the models were also trained without regular term in order to justify whether the regular term is necessary. The results of ablation experiments are reported in the following Figure \ref{fig:ablation1} and \ref{fig:ablation2}.

\begin{figure*}[!h]
\centering
    \includegraphics[width=0.9\linewidth]{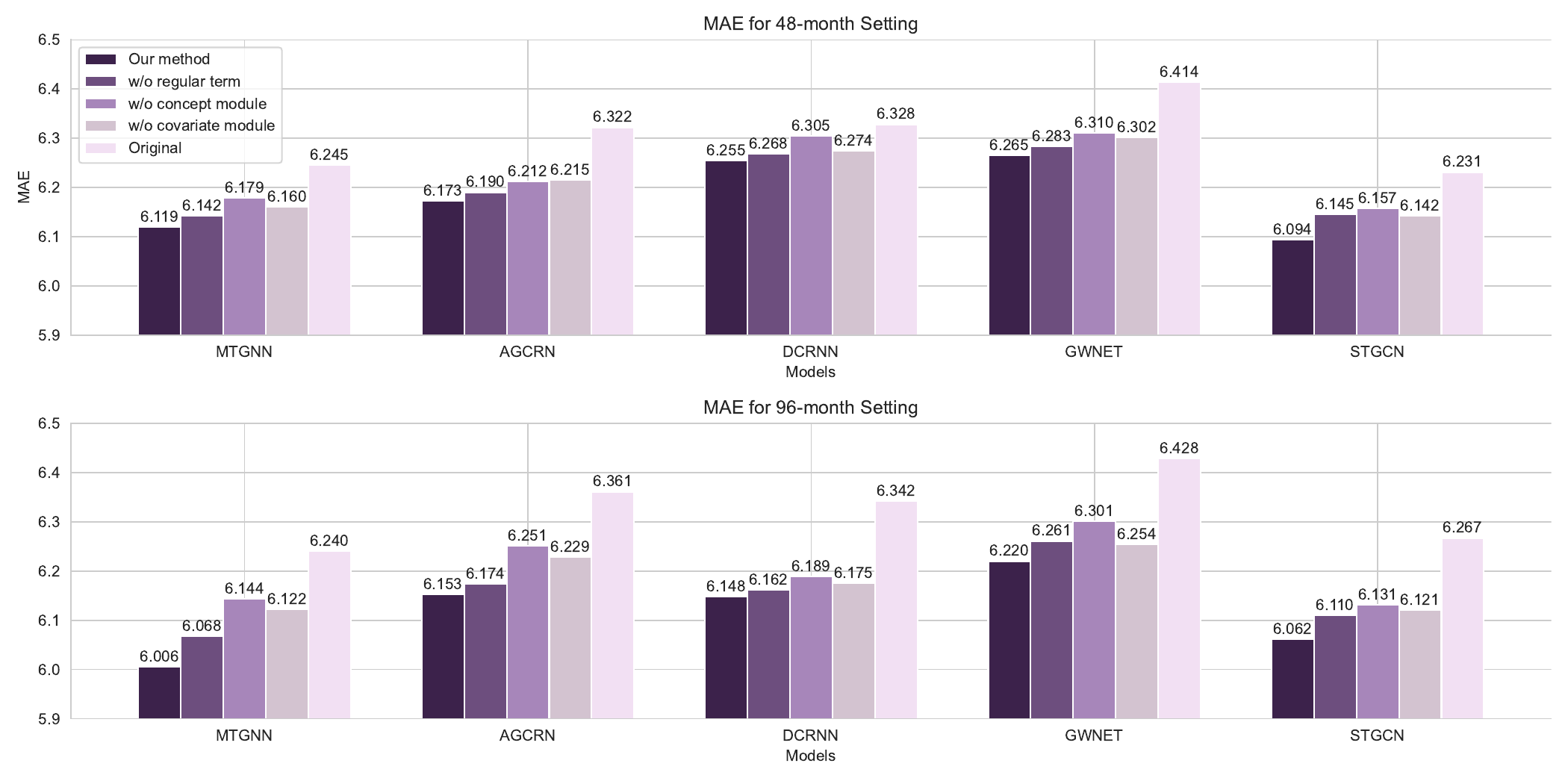}
    \caption{Ablation experiment result of NYCBIKE dataset}
    \label{fig:ablation1}
\end{figure*}
\begin{figure*}[!h]
\centering
    \includegraphics[width=0.9\linewidth]{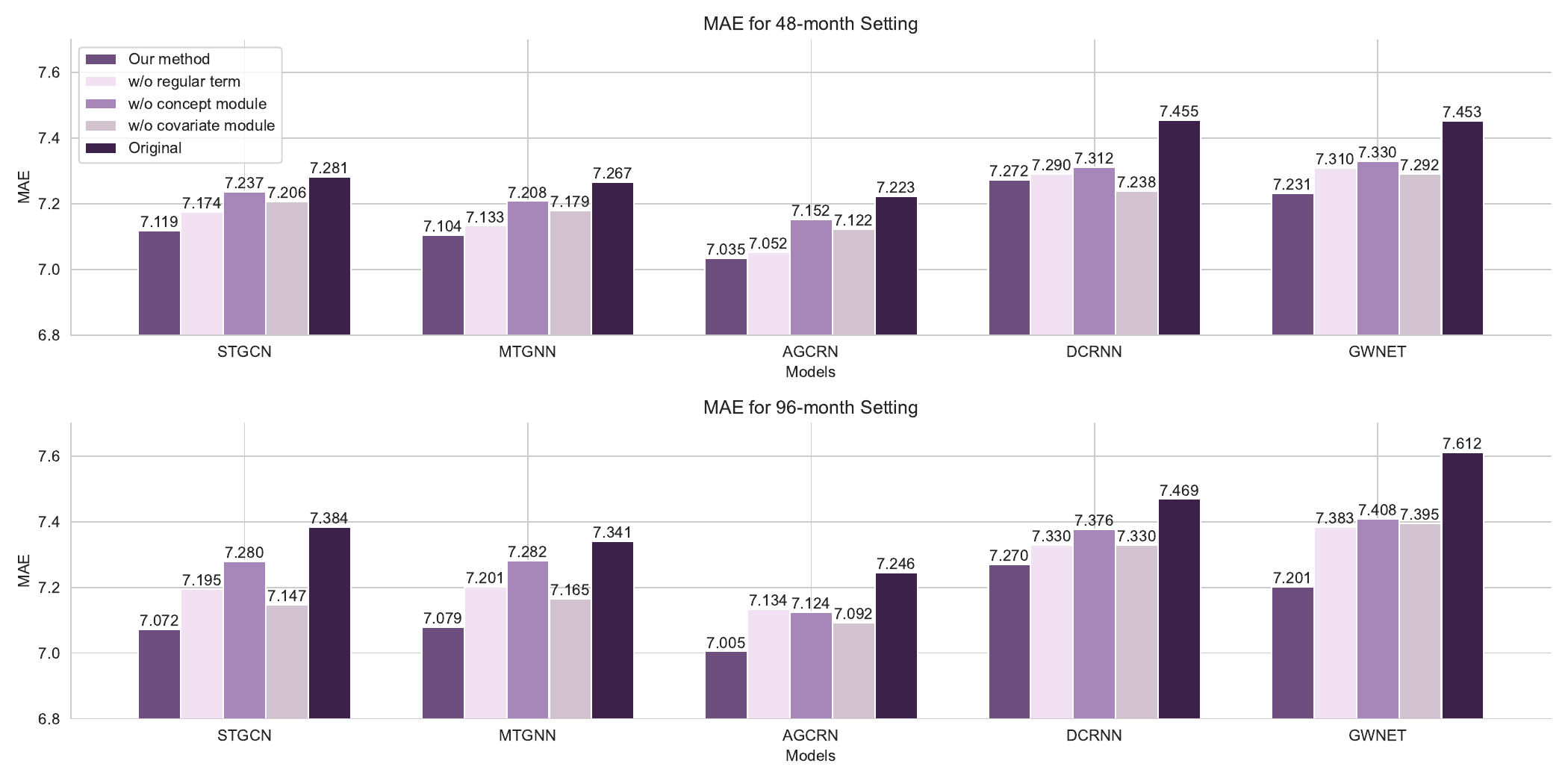}
    \caption{Ablation experiment result of NYCTAXI dataset}
    \label{fig:ablation2}
\end{figure*}

It can be observed from ablation experiments that the performance will decrease without each part of our method, which can vindicate the effectiveness of our modules. Besides the benefit of regular term can be also proved by the ablation experiments.

Moreover, the efficacy of concept shift module is often more significant than the covariate shift module. For example, when using NYCTAXI data spanning 96 months to train a STGCN model, our method yields the MAE in 7.072, but if the model is trained without concept shift module, the MAE will increase to 7.280, however, the MAE only increase to 7.147 when training without covariate shift module. And this trend can also be observed in other cases.

This result implies that the pattern change in bike and taxi usage is more significant than the magnitude change. Therefore, just aligning the statistics of covariate (RevIN) or features of covariate (AdaRNN) cannot obtain a satisfied prediction accuracy.

\subsection{Result analysis}

\begin{figure*}[!h]
    \centering
    \includegraphics[width=0.9\linewidth]{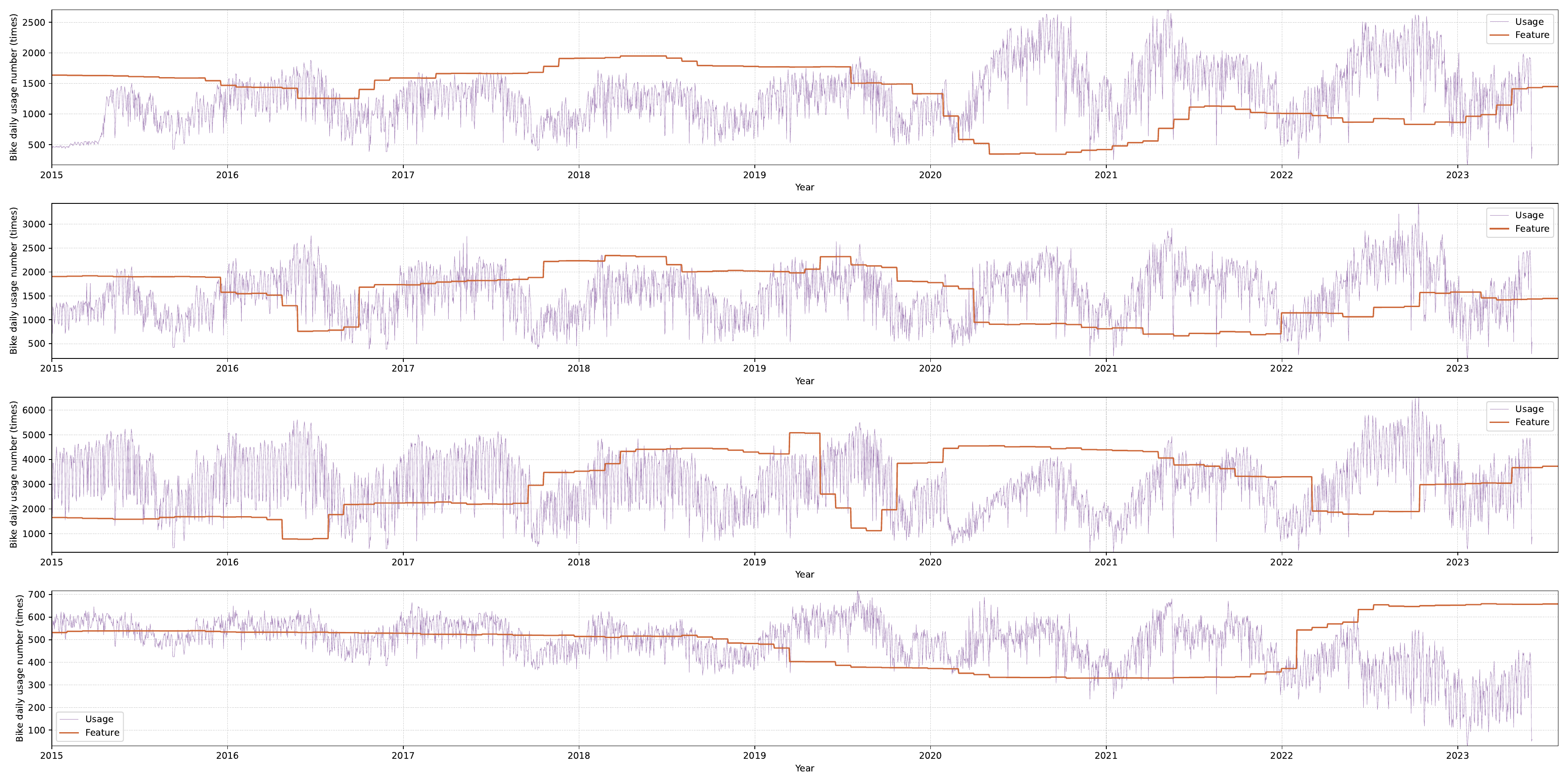}
    \caption{The usage numbers and environment features in different grids for NYCBIKE dataset}
    \label{fig:bike_feature}
\end{figure*}

Even though the performances of our method are significant, we still hope to give some explanation of our methods. In another word, we want to answer whether there are some actual meanings of the environment features learned by our method. Therefore, we plot the bike usage in some regions and the environment features learned by our method for these regions, as the Figure \ref{fig:bike_feature} shows.

From the Figure \ref{fig:bike_feature}, it can be observed that the change of the environment features learned by our method is consistent with the change of bike usage. For example, in the top subfigure, there is a distinct changing point of environment feature in the beginning of 2020. And this changing point can also be observed in the bike usage numbers. For the other three subplots, the change points in 2020 and 2022 can also be find for both environment features and bike usage patterns. Therefore, it can be concluded that the environment features learned by our method can reflects the bike usage patterns in some extents.
\begin{figure*}[!h]
    \centering
    \includegraphics[width=0.9\linewidth]{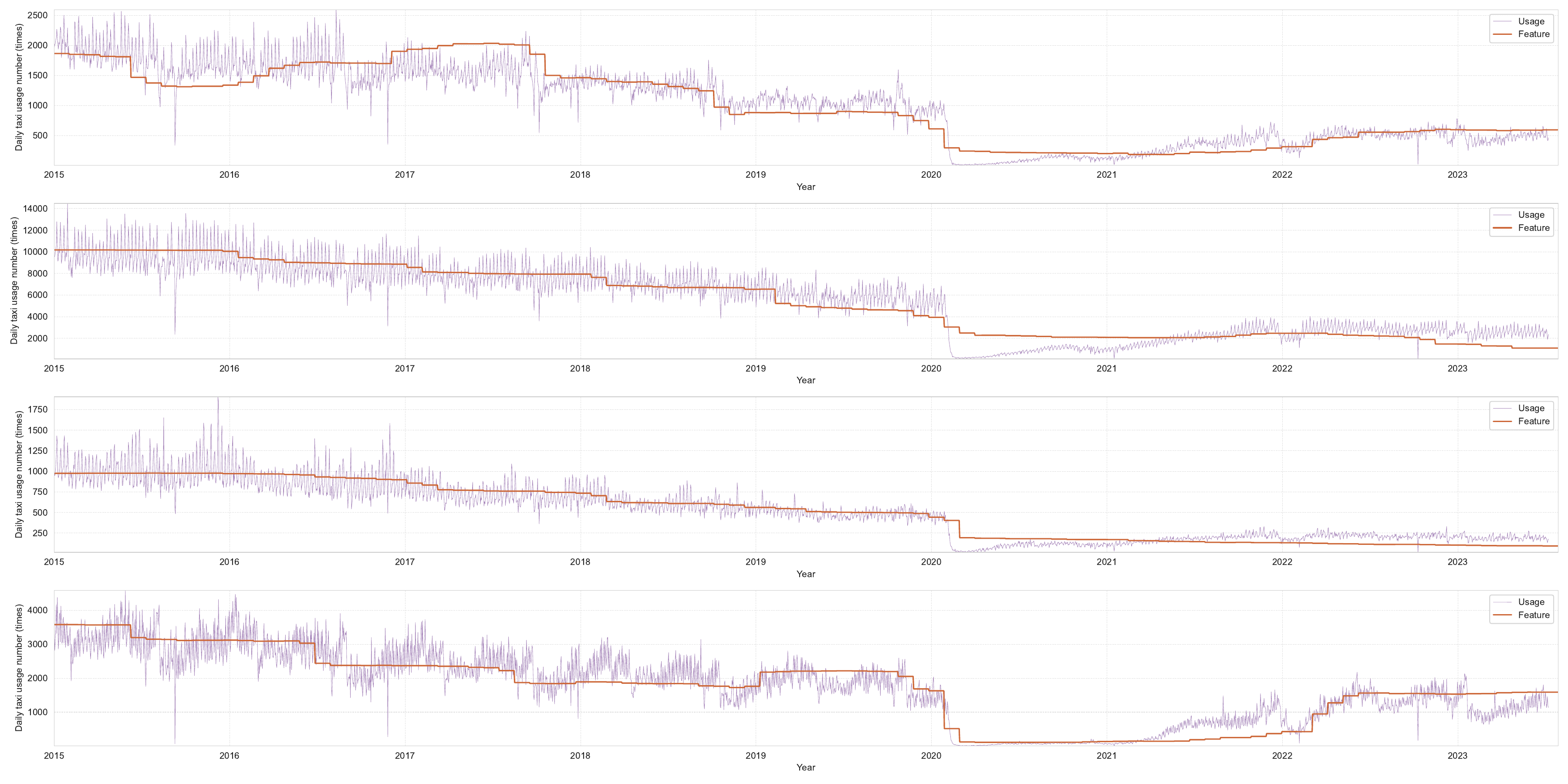}
    \caption{The usage numbers and environment features in different regions for NYCTAXI dataset}
    \label{fig:taxi_feature}
\end{figure*}
Besides we also plot the daily taxi usage number and the environment features for taxi dataset in the Figure \ref{fig:taxi_feature}. It can be concluded that the taxi usage pattern has a very distinctive change point at around 2020 which might because of the COVID-19. And the environment features also see a significant change point at 2020, so we can also conclude that the environment features learned for the taxi data set can also reflect the patterns of taxi usage.
\section{Conclusion}
In this research, we focus on two main questions, the first is that what result will be if we use a larger dataset to train the traffic prediction model, and the second question is how to make good use of more historical data to calibrate traffic prediction models. 

For the first question, two large data set were collected and used to calibrate 5 popular traffic prediction models and the results indicated that larger datasets cannot ensure better performance when testing. Then we focused on the reasons behind this contradictory observation and concluded that the distribution shift may be the reason. Afterwards, a careful analysis was conducted and both covariance shift and concept shift were found. Therefore, in order to use the larger dataset more effectively, different modules were devised to address these different shifts. Environment aware learning was deployed to address conception shift. Specifically, we proposed an environment feature pool to model different transportation patterns in different periods. Besides, adjacent matrix generating models was proposed to address the changing spatial relationships in different regions. Finally, to tackle covariate shift, we use a sample weighting method considering both similarly and recency.

The experimental results demonstrate that models trained by our method exhibit improved accuracy compared to original models and some baselines. Furthermore, with the accumulation of historical data, models trained with our method demonstrate a continuous increase in accuracy when testing. Besides, some ablation experiments were also conducted to demonstrate the effectiveness of all components in our method and the analysis of environmental features proved that our method could learn the changing traffic patterns in some extent.

In summary, to the best of our knowledge, this study represents the first attempt to calibrate traffic prediction models using an extensive eight-year dataset. In practical scenarios, the transportation system will amass a substantial amount of historical data over time. Our study underscores the necessity for specialized designs to efficiently leverage this data, emphasizing it as a topic deserving further research.

\bibliographystyle{IEEEtran}
\bibliography{ref}


%



%
 \begin{IEEEbiography}
     [{\includegraphics[width=1in,height=1.25in,clip,keepaspectratio]{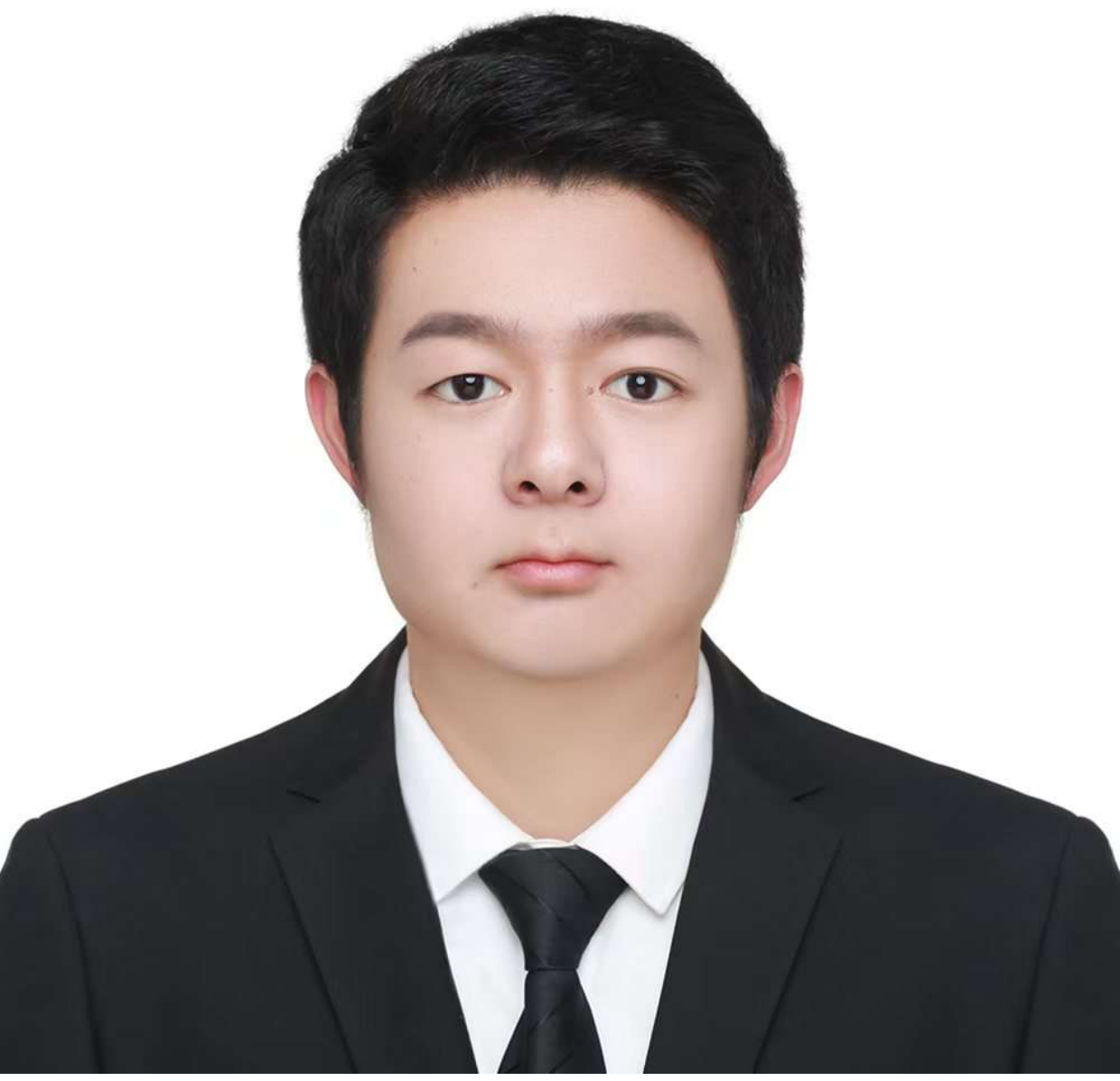}}]{Xiannan Huang} 
   received Bachelor's degree in civil engineering from Tongji University, Shanghai, and currently pursuing a Ph.D. degree at the School of Transportation Engineering, Tongji University. His research interests include spatial-temporal data analysis and trip prediction.
\end{IEEEbiography}
\begin{IEEEbiography}
    [{\includegraphics[width=1in,height=1.25in,clip,keepaspectratio]{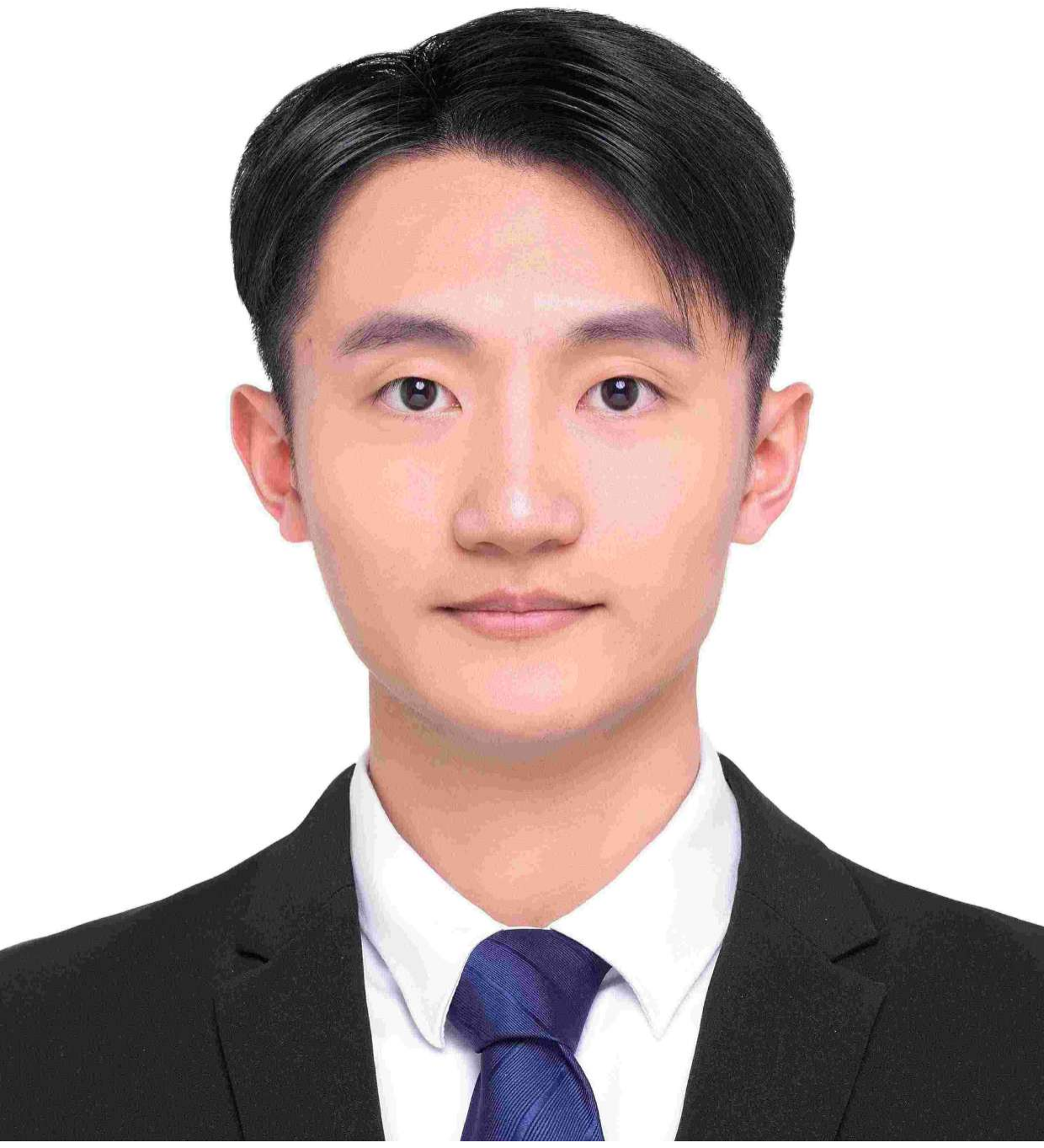}}]{
Shuhan Qiu} received Bachelor's degree in transportation engineering from Tongji University, Shanghai, and currently pursuing a Ph.D. degree at the School of Transportation Engineering, Tongji University. His research interests include route choice model, path reconstruction, deep learning in transportation system and automatic driving.
\end{IEEEbiography}
\begin{IEEEbiography}
    [{\includegraphics[width=1in,height=1.25in,clip,keepaspectratio]{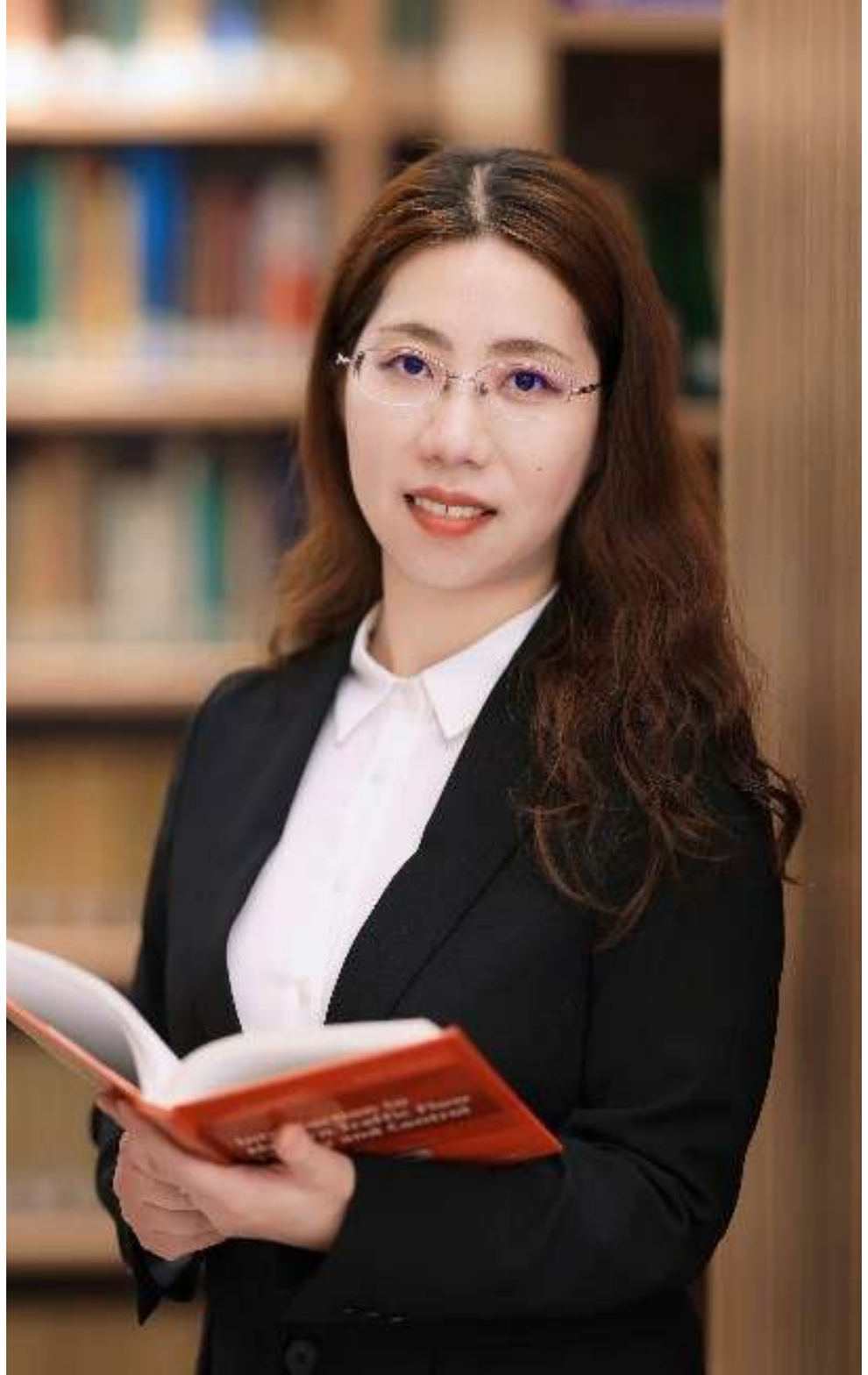}}]{
Yan Cheng} received the B.S. degree in transportation engineering and the Ph.D. degree in road and railway engineering from Tongji University, China, in 2012 and 2018, respectively. She is currently a senior researcher and doctoral supervisor in the College of Transportation Engineering, Tongji University. She takes the planning and design of intelligent rail transit network as her research direction, takes the theory and method of rail transit passenger flow prediction as her starting point, devotes herself to promoting the application of artificial intelligence in intelligent rail transit system.
\end{IEEEbiography}

\begin{IEEEbiography}[{\includegraphics[width=1in,height=1.25in,clip,keepaspectratio]{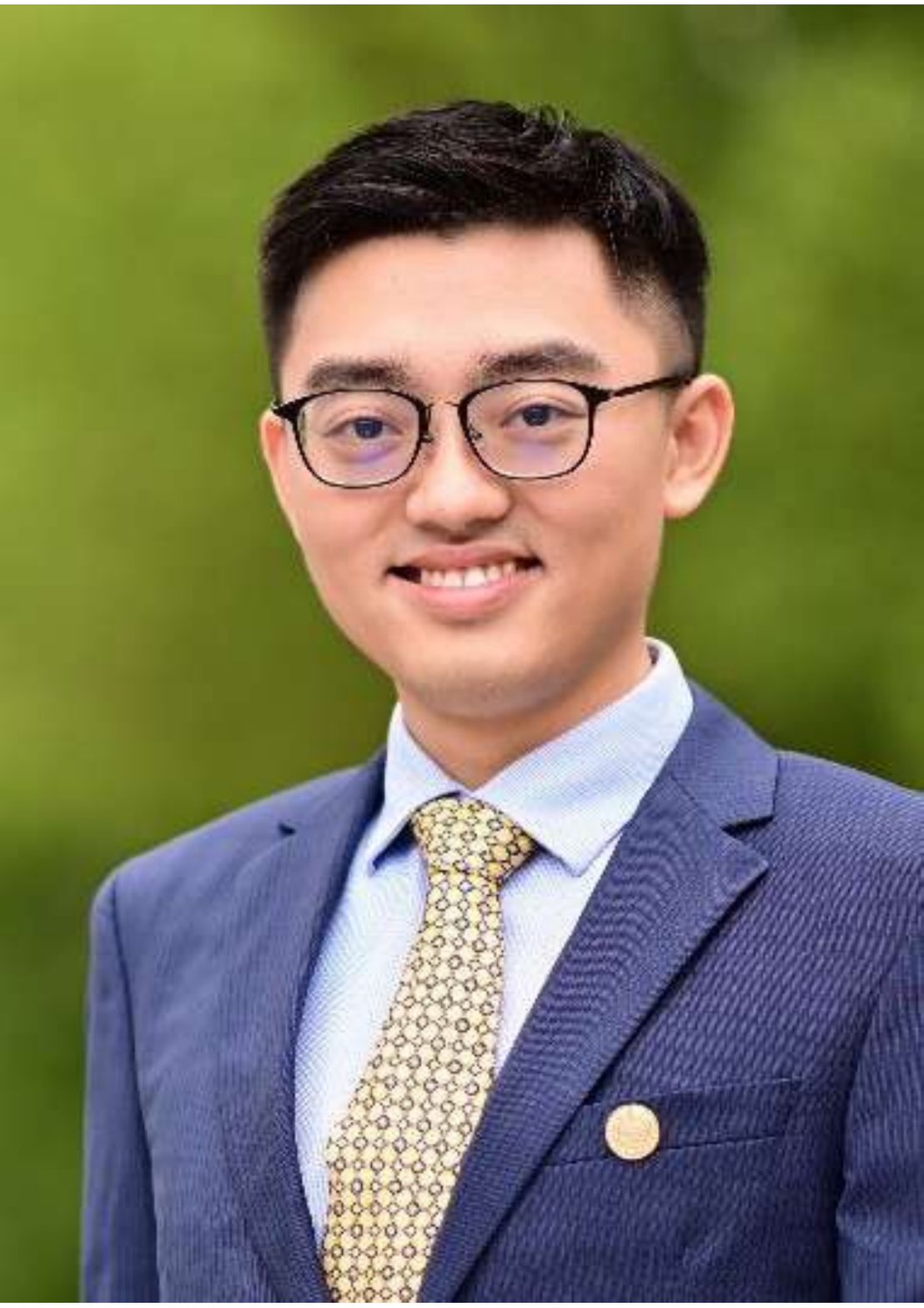}}]{Quan Yuan}
received B.S. degree in urban planning and Economics from Peking University in 2011 and received a Master's degree in urban and Regional planning from UCLA in 2013. He  a Ph.D. degree in urban planning and development from University of Southern California in 2018. He is now a senior researcher and doctoral supervisor at Urban Mobility Institute of Tongji University. His current research interests are transport-environment-policy research, urban freight, traffic big data mining  artificial intelligence, urban transportation planning and urban parking

\end{IEEEbiography}
\begin{IEEEbiography}[{\includegraphics[width=1in,height=1.25in,clip,keepaspectratio]{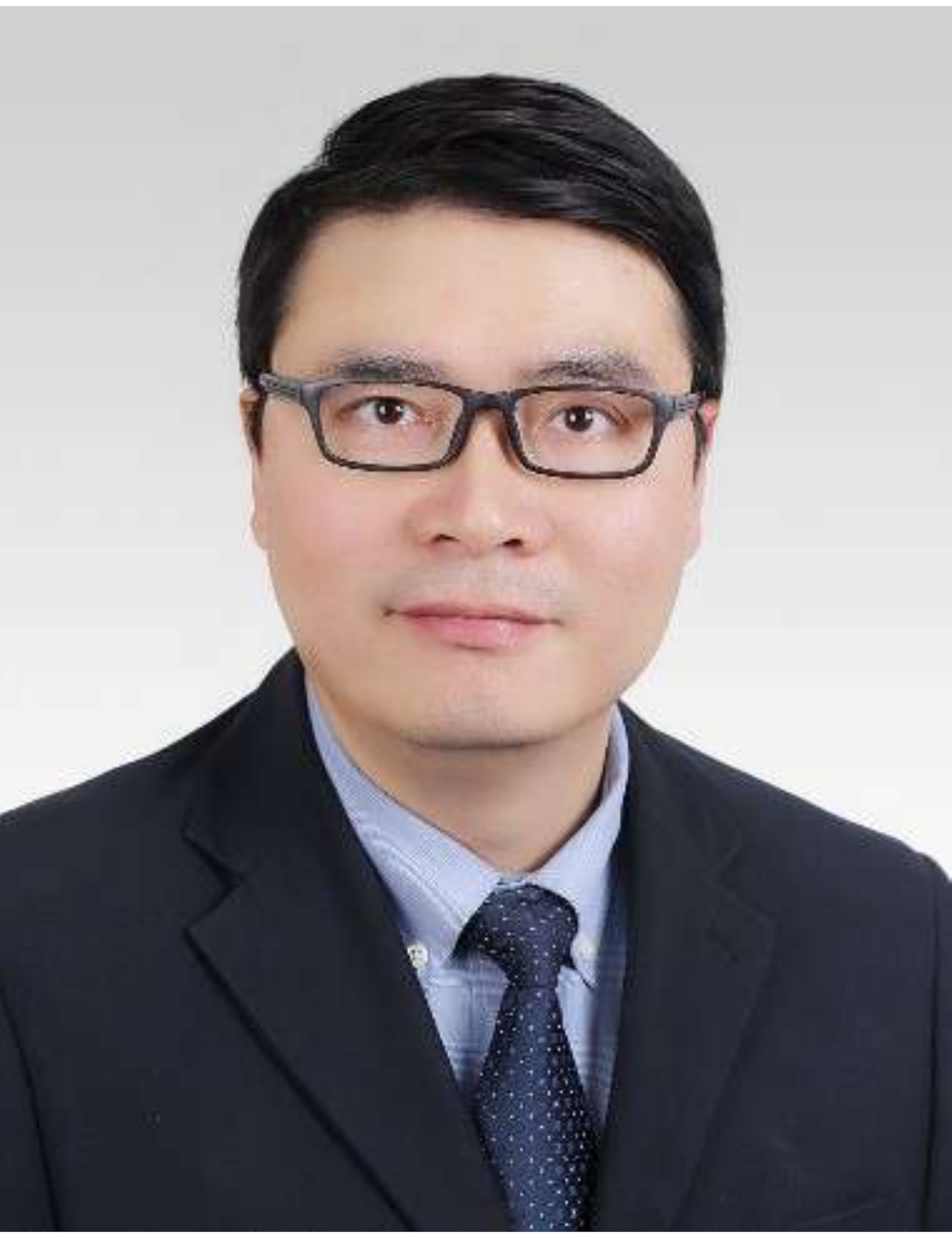}}]{Chao Yang}
 received B.S. degree in Transportation Engineering from Tongji University and Ph.D. degree in Municipal Engineering from Tongji University in 1994 and 1999 respectively, and is currently a professor and doctoral supervisor in the College of Transportation Engineering of Tongji University. He presided 3 National Natural Science fundation of China and has published about 200 papers in international academic journals in the field of transportation research. His research interests include transportation big data analysis, bus trip behavior analysis, activity-based models, and transportation planning theory.
\end{IEEEbiography}



\end{document}